\documentclass[10pt,journal,compsoc]{IEEEtran}

\usepackage{booktabs}
\usepackage{xcolor}
\usepackage{graphicx}
\usepackage{multirow}
\usepackage{url}
\usepackage[numbers]{natbib}

\usepackage{tikz}
\usepackage[edges]{forest}
\definecolor{hiddendraw}{RGB}{205, 44, 36}
\definecolor{hidden-blue}{RGB}{194,232,247}
\definecolor{hidden-orange}{RGB}{243,202,120}
\definecolor{hidden-yellow}{RGB}{242,244,193}

\ifCLASSOPTIONcompsoc
  \usepackage[nocompress]{cite}
\else
  \usepackage{cite}
\fi

\begin{document}

\title{A Survey on Aspect-Based Sentiment Analysis: Tasks, Methods, and Challenges}

\author{Wenxuan~Zhang, Xin~Li, Yang~Deng, Lidong~Bing, and~Wai~Lam
\IEEEcompsocitemizethanks{\IEEEcompsocthanksitem Wenxuan Zhang, Xin Li and Lidong Bing are with DAMO Academy, Alibaba Group. \protect\\
E-mail: \{saike.zwx, xinting.lx, l.bing\}@alibaba-inc.com
\IEEEcompsocthanksitem Yang Deng and Wai Lam are with The Chinese Univerisity of Hong Kong. \protect\\
E-mail: \{ydeng, wlam\}@se.cuhk.edu.hk}
\thanks{This work has been submitted to the IEEE for possible publication. Copyright may be transferred without notice, after which this version may no longer be accessible.}}

\IEEEtitleabstractindextext{%
\begin{abstract}
As an important fine-grained sentiment analysis problem, aspect-based sentiment analysis (ABSA), aiming to analyze and understand people's opinions at the aspect level, has been attracting considerable interest in the last decade. To handle ABSA in different scenarios, various tasks are introduced for analyzing different sentiment elements and their relations, including the aspect term, aspect category, opinion term, and sentiment polarity. Unlike early ABSA works focusing on a single sentiment element, many compound ABSA tasks involving multiple elements have been studied in recent years for capturing more complete aspect-level sentiment information. However, a systematic review of various ABSA tasks and their corresponding solutions is still lacking, which we aim to fill in this survey. More specifically, we provide a new taxonomy for ABSA which organizes existing studies from the axes of concerned sentiment elements, with an emphasis on recent advances of compound ABSA tasks. From the perspective of solutions, we summarize the utilization of pre-trained language models for ABSA, which improved the performance of ABSA to a new stage. Besides, techniques for building more practical ABSA systems in cross-domain/lingual scenarios are discussed. Finally, we review some emerging topics and discuss some open challenges to outlook potential future directions of ABSA.
\end{abstract}

}

\maketitle

\IEEEdisplaynontitleabstractindextext

\IEEEpeerreviewmaketitle

\ifCLASSOPTIONcompsoc
\IEEEraisesectionheading{\section{Introduction}\label{sec:introduction}}
\else
\section{Introduction}
\label{sec:introduction}
\fi
\IEEEPARstart{D}{iscovering} and understanding opinions from online user-generated content is crucial for widespread applications. For example, analyzing customer sentiments and opinions from reviews in E-commerce platforms helps improve the product or service, and make better marketing campaigns. Given the massive amount of textual content, it is intractable to manually digest the opinion information. Therefore, designing an automatic computational framework for analyzing opinions hidden behind the unstructured texts is necessary, resulting in the emergence of the research field \textit{\textbf{sentiment analysis and opinion mining}}~\citep{hlt12-liu-bing-sa}. 

Conventional sentiment analysis studies mainly perform prediction at the sentence or document level \citep{acl02-thumbs, emnlp02-pang-thumbs, emnlp02-yu-sa}, identifying the overall sentiment towards the whole sentence or document. To make the prediction, it is assumed that a single sentiment is conveyed towards the single topic in the given text, which may not be the case in practice. Under this circumstance, the need for recognizing more fine-grained aspect-level opinions and sentiments, dubbed as \textit{\textbf{Aspect-Based Sentiment Analysis}} (ABSA), has received increasing attention in the past decade~\citep{tkde16-absa-survey, tac20-absa-survey}. In the ABSA problem, the concerned target on which the sentiment is expressed shifts from an entire sentence or document to an entity or a certain aspect of an entity. For instance, an entity can be a specific product in the E-commerce domain, and its property or characteristics such as the price and size are the aspects of it. Since an entity can also be regarded as a special ``general'' aspect, we collectively refer to an entity and its aspect as ``aspect'' in this paper. ABSA is thus the process of building a comprehensive opinion summary at the aspect level, which provides useful fine-grained sentiment information for downstream applications.

\begin{figure}[!t]
\centering
\includegraphics[width=\linewidth]{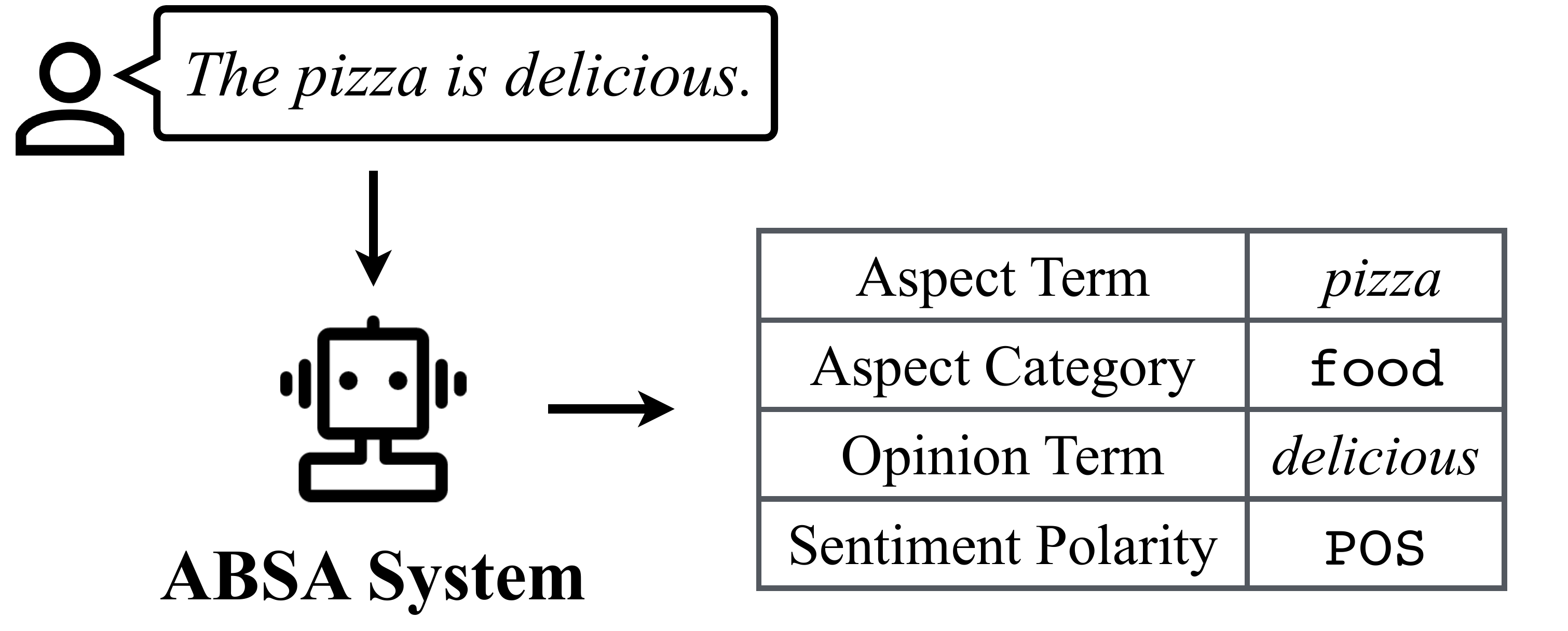}
\caption{An example of the four key sentiment elements of ABSA.} 
\label{fig:absa-overview}
\end{figure}

Generally, the main research line of ABSA involves the identification of various aspect-level sentiment elements, namely, aspect terms, aspect categories, opinion terms and sentiment polarities \citep{emnlp21-quad}. As shown in Fig.~\ref{fig:absa-overview}, given a sentence ``\textit{The pizza is delicious.}'', the corresponding sentiment elements are ``\textit{pizza}'', ``\verb|food|'', ``\textit{delicious}'', and ``\verb|positive|'' respectively, where ``\textit{pizza}'' and ``\textit{delicious}'' are explicitly expressed, ``\verb|food|'' and ``\verb|positive|'' belong to the pre-defined category and sentiment sets. 
Early works in ABSA begin with identifying each single sentiment element separately. For instance, the aspect term extraction (ATE) task \citep{emnlp15-ate-rnn} aims to extract all mentioned aspect terms in the given text; while the aspect sentiment classification task \citep{acl11-asc-jiang} predicts the sentiment polarity for a specific aspect within a sentence. In this paper, we refer to these tasks as \textbf{Single ABSA} tasks.

However, finding a single sentiment element is still far from satisfactory for understanding more complete aspect-level opinion, which requires not only the extraction of multiple sentiment elements but also the recognition of the correspondence and dependency between them. To this end, several new ABSA tasks \citep{emnlp13-uabsa, acl20-aope-sync, aaai20-aste, emnlp21-quad} together with corresponding benchmark datasets have been introduced in recent years to facilitate the study on the joint prediction of multiple sentiment elements. These tasks are referred to as \textbf{Compound ABSA} tasks, in contrast to Single ABSA tasks involving a single sentiment element only. 
For example, the aspect-opinion pair (AOPE) extraction task \citep{acl20-aope-sync, acl20-aope-spanmlt} requires extracting the aspect and its associated opinion term in a compound form, i.e., extracting (\textit{pizza}, \textit{delicious}) pair from the previous example sentence. It thus provides a clearer understanding of what the mentioned opinion target and its associated opinion expression are.
Following some pioneering works, a wide variety of frameworks have been proposed to tackle different compound ABSA tasks for enabling aspect-level opinion mining in different scenarios. 
However, a systematic review of various ABSA tasks, especially recent progress on compound ABSA tasks, is lacking in the existing surveys \citep{hlt12-liu-bing-sa, tkde16-absa-survey, access19-absa-survey, tac20-absa-survey, tcss20-absa-survey, tac20-sa-survey}, which we aim to fill through this survey paper.

Aside from designing specific models for different tasks, the advent of pre-trained language models (PLMs) such as BERT  \citep{naacl19-bert} and RoBERTa \citep{arxiv19-roberta} has brought substantial improvements on a wide range of ABSA tasks in recent years. With PLMs as the backbone, the generalization capability and the robustness of ABSA models have been significantly improved. For example, \citet{wnut19-exploiting} show that using a simple linear classification layer stacked on top of BERT can achieve more competitive performance than previous specifically designed neural models for the End-to-End ABSA task. Although constructing ABSA models based on PLMs has become ubiquitous nowadays, they are not discussed in the existing surveys \citep{tac20-absa-survey,tcss20-absa-survey,tac20-sa-survey} due to their recency of publication. Therefore, in this paper, we provide an in-depth analysis of existing PLM-based ABSA models by discussing both their advances and limitations.

To conduct ABSA in practical settings, we also review the recent works tackling the cross-domain and cross-lingual ABSA problem.
Current ABSA models that achieved satisfactory performance in various tasks often hold a common assumption: the training and testing data come from the same distribution (e.g., the same domain or the same language). When the distribution of data changes, re-training the ABSA model is often needed to guarantee the performance. However, it is usually expensive or even impossible to collect additional large volume of labeled data, especially for the ABSA task requiring aspect-level annotations. In this case, adapting the trained model to unseen domains, i.e., cross-domain transfer~\citep{tkde10-cross-domain-survey} or unseen languages, i.e., cross-lingual transfer~\citep{jair19-cross-lingual-survey}, provides an alternative solution for building ABSA systems well generalizing to different domains and languages.

There have been other surveys and reviews about ABSA. Existing surveys of general sentiment analysis research \citep{hlt12-liu-bing-sa, tac20-sa-survey} discuss the ABSA problem, but they do not provide a detailed description of the recent advances and challenges. The earliest ABSA survey by \citet{tkde16-absa-survey} comprehensively introduces the ABSA studies before 2016, but it mainly focuses on the non-neural ABSA methods. \citet {access19-absa-survey}, \citet{tcss20-absa-survey}, and \citet{tac20-absa-survey} introduce the deep learning based ABSA models. However, their discussions are only limited to single ABSA tasks with a few pioneering works on the End-to-End ABSA task. A comprehensive review of all ABSA tasks, the impacts of PLMs for the ABSA problem, as well as recent progress on the cross-domain/lingual transfer are not covered. 

Overall, the main goal of this survey paper is to systematically review the advances and challenges of the ABSA problem from a modern perspective. More specifically, we provide a new taxonomy for ABSA which organizes various ABSA studies from the axes of concerned sentiment elements, with an emphasis on the compound ABSA tasks studied in recent years. Along this direction, we discuss and summarize all kinds of methods proposed for each task. Besides, we investigate the potentials and limitations of exploiting pre-trained language models for the ABSA problem. We also summarize the research efforts on cross-domain and cross-lingual ABSA. Finally, we discuss some emerging trends and open challenges, aiming to shed light on potential future directions in this area. We maintain a GitHub repository to collect useful resources such as the paper list discussed in this survey, links to tools and datasets: {\ttfamily \url{https://github.com/IsakZhang/ABSA-Survey}}.

\section{Background}

\tikzstyle{mybox}=[
    rectangle,
    draw=hiddendraw,
    rounded corners,
    text opacity=1,
    minimum height=2em,
    minimum width=15em,
    inner sep=2pt,
    align=center,
    fill opacity=.5,
    ]
    
\tikzstyle{leaf}=[mybox,minimum height=1em,
fill=hidden-blue!40, text width=20em,  text=black,align=left,font=\scriptsize,
inner xsep=2pt,
inner ysep=1pt,
]

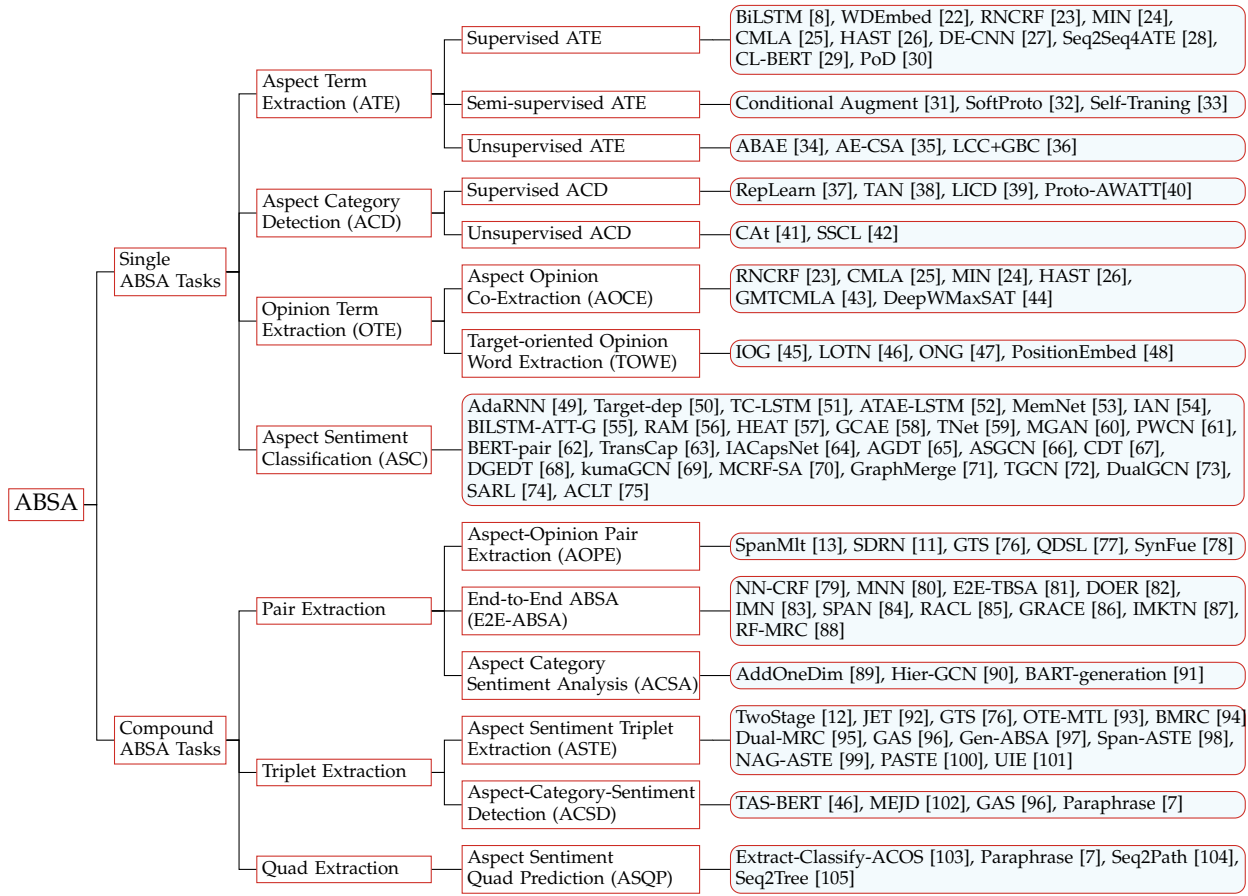
\begin{figure*}[tp]
  \centering
  \begin{forest}
    forked edges,
    for tree={
      grow=east,
      reversed=true,
      anchor=base west,
      parent anchor=east,
      child anchor=west,
      base=left,
      font=\small,
      rectangle,
      draw=hiddendraw,
      align=left,
      minimum width=2.5em,
      s sep=6pt,
      inner xsep=2pt,
      inner ysep=1pt,
      ver/.style={rotate=90, child anchor=north, parent anchor=south, anchor=center},
    },
    where level=1{text width=4em,font=\scriptsize}{},
    where level=2{text width=6.5em,font=\scriptsize}{},
    where level=3{text width=9em,font=\scriptsize}{},
    [ABSA
    [Single\\ABSA Tasks
        [Aspect Term\\Extraction (ATE)
           [Supervised ATE
                [BiLSTM~\citep{emnlp15-ate-rnn}{,} WDEmbed~\citep{ijcai16-ate-wordembed}{,} RNCRF~\citep{emnlp16-ate-rncrf}{,} MIN~\citep{emnlp17-ate-min}{,}\\ CMLA~\citep{aaai17-ate-cmla}{,} HAST~\citep{ijcai18-ate-hast}{,} DE-CNN~\citep{acl18-ate-decnn}{,} Seq2Seq4ATE~\citep{acl19-ate-seq2label}{,}\\CL-BERT~\citep{coling20-ate-constituency}{,} PoD~\citep{coling20-ate-pod}
            ,leaf]
           ]
           [Semi-supervised ATE
                [Conditional Augment~\citep{acl20-ate-maskedseq2seq}{,} SoftProto~\citep{emnlp20-ate-softproto}{,} Self-Traning~\citep{emnlp21-ate-selftrain}
            ,leaf]
           ]
           [Unsupervised ATE
               [ABAE~\citep{acl17-ruidan-ate}{,} AE-CSA~\citep{ijcai19-unsupervised-ate}{,} LCC+GBC~\citep{emnlp19-ate-coupling}
               ,leaf]
           ]
        ]
        [Aspect Category\\Detection (ACD)
            [Supervised ACD
                [RepLearn~\citep{aaai15-acd-zhou}{,} TAN~\citep{arxiv19-acd-topic}{,} LICD~\citep{ecir19-acd-matching}{,} Proto-AWATT\citep{acl21-acd-multi-label}
                ,leaf]
            ]
            [Unsupervised ACD
                [CAt~\citep{acl20-acd-cat}{,} SSCL~\citep{aaai21-acd}
                ,leaf]
            ]
        ]
        [Opinion Term\\Extraction (OTE)
            [Aspect Opinion \\Co-Extraction (AOCE)
                [RNCRF~\citep{emnlp16-ate-rncrf}{,} CMLA~\citep{aaai17-ate-cmla}{,} MIN~\citep{emnlp17-ate-min}{,}  HAST~\citep{ijcai18-ate-hast}{,} \\ GMTCMLA~\citep{taslp19-aoce-yu}{,}  DeepWMaxSAT~\citep{emnlp20-aoce-wwy}
                ,leaf]
            ]
            [Target-oriented Opinion\\Word Extraction (TOWE)
                [IOG~\citep{naacl19-towe}{,} LOTN~\citep{aaai20-tasd}{,} ONG~\citep{emnlp20-towe-syntax}{,} PositionEmbed~\citep{emnlp21-towe-position}
                ,leaf]
            ]
        ]
        [Aspect Sentiment\\Classification (ASC)
            [AdaRNN~\citep{acl14-asc-dong}{,} Target-dep~\citep{ijcai15-asc-nn}{,} TC-LSTM~\citep{coling16-asc-effective-lstm}{,} ATAE-LSTM~\citep{emnlp16-asc-atae}{,} MemNet~\citep{emnlp16-asc-memory}{,} IAN~\citep{ijcai17-asc-ian}{,}\\BILSTM-ATT-G~\citep{eacl17-asc-attention}{,} RAM~\citep{emnlp17-asc-ram}{,} HEAT~\citep{cikm17-asc-heat}{,} GCAE~\citep{acl18-asc-cnn}{,} TNet~\citep{acl18-asc-lx-tnet}{,} MGAN~\citep{emnlp18-asc-mgan}{,} PWCN~\citep{sigir19-asc-pwcn}{,}\\
            BERT-pair~\citep{naacl19-bertabsa}{,}
            TransCap~\citep{acl19-asc-transcap}{,} IACapsNet~\citep{emnlp19-asc-capsule}{,} AGDT~\citep{emnlp19-asc-agdt}{,} ASGCN~\citep{emnlp19-asgcn}{,} CDT~\citep{emnlp19-asc-cdt}{,}\\ DGEDT~\citep{acl20-asc-dgedt}{,}
            kumaGCN~\citep{emnlp20-latent-graph}{,} MCRF-SA~\citep{emnlp20-asc-xulu}{,} GraphMerge~\citep{naacl21-asc-graphmerge}{,} TGCN~\citep{naacl21-tgcn}{,} DualGCN~\citep{acl21-asc-dualgcn}{,}\\ SARL~\citep{emnlp21-asc-eliminating}{,} ACLT~\citep{emnlp21-asc-aclt}
            ,leaf,text width=30.6em]
        ]
    ]
    [Compound \\ ABSA Tasks
        [Pair Extraction
            [Aspect-Opinion Pair\\Extraction (AOPE)         
                [SpanMlt~\citep{acl20-aope-spanmlt}{,} SDRN~\citep{acl20-aope-sync}{,} GTS~\citep{emnlp20-aste-grid}{,} QDSL~\citep{aaai21-aope-mrc}{,} SynFue~\citep{ijcai21-aope-syntax}
                ,leaf]
            ]
            [End-to-End ABSA\\ (E2E-ABSA)         
                [NN-CRF~\citep{emnlp15-uabsa}{,} MNN~\citep{ijcnn18-uabsa}{,} E2E-TBSA~\citep{aaai19-uabsa-lx}{,} DOER~\citep{acl19-uabsa-doer}{,}\\IMN~\citep{acl19-uabsa-imn}{,} SPAN~\citep{acl19-openabsa}{,} RACL~\citep{acl20-uabsa-racl}{,}  GRACE~\citep{emnlp20-uabsa-grace}{,} IMKTN~\citep{emnlp21-uabsa-iterative}{,}\\RF-MRC~\citep{emnlp21-uabsa-self}
                ,leaf]
            ]
            [Aspect Category\\Sentiment Analysis (ACSA)         
                [AddOneDim~\citep{emnlp18-acsa-joint}{,}  Hier-GCN~\citep{coling20-acsa-hiergcn}{,} BART-generation~\citep{emnlp21-acsa-seq2seq}
                ,leaf]
            ]
        ]
        [Triplet Extraction
            [Aspect Sentiment Triplet\\ Extraction (ASTE)
                [TwoStage~\citep{aaai20-aste}{,} JET~\citep{emnlp20-aste-position}{,} GTS~\citep{emnlp20-aste-grid}{,} OTE-MTL~\citep{emnlp20-aste-mtl}{,} BMRC~\citep{aaai21-aste-bimrc}\\Dual-MRC~\citep{aaai21-aste-dualmrc}{,} GAS~\citep{acl21-gabsa}{,} Gen-ABSA~\citep{acl21-aste-generative-qxp}{,} Span-ASTE~\citep{acl21-aste-span}{,}\\NAG-ASTE~\citep{tnnls21-aste-nonautoregressive}{,} PASTE~\citep{emnlp21-aste-pointer}{,} UIE~\citep{acl22-uie}
                ,leaf]
                ]
            [Aspect-Category-Sentiment\\ Detection (ACSD)
                [TAS-BERT~\citep{aaai20-tasd}{,} MEJD~\citep{kbs21-tasd-mejd}{,} GAS~\citep{acl21-gabsa}{,} Paraphrase~\citep{emnlp21-quad}
                ,leaf]
                ]
        ]
        [Quad Extraction
            [Aspect Sentiment\\Quad Prediction (ASQP)
                [Extract-Classify-ACOS~\citep{acl21-quad-cai}{,} Paraphrase~\citep{emnlp21-quad}{,} Seq2Path~\citep{acl22-seq2path}{,}\\Seq2Tree~\citep{ijcai22-opinion-tree}
                ,leaf]
                ]
        ]
    ]
    ]
  \end{forest}
  \caption{Taxonomy of ABSA tasks, with representative methods of each task. }
  \label{fig:absa-tasks}
\end{figure*}

\subsection{Four Sentiment Elements of ABSA} \label{sec:four-elements}
According to \citet{hlt12-liu-bing-sa}, the general sentiment analysis problem consists of two key components: target and sentiment. 
For ABSA, the target can be described with either an aspect category $c$ or an aspect term $a$, while the sentiment involves a detailed opinion expression - the opinion term $o$, and a general sentiment orientation - the sentiment polarity $p$. These four sentiment elements constitute the main line of ABSA research: 
\begin{itemize}
    \item \textbf{aspect category} $c$ defines a unique aspect of an entity and is supposed to fall into a category set $\mathcal{C}$, pre-defined for each specific domain of interest. For example, \verb|food| and \verb|service| can be aspect categories for the \textit{restaurant} domain. 
    \item \textbf{aspect term} $a$ is the opinion target which explicitly appears in the given text, e.g., ``\textit{pizza}'' in the sentence ``\textit{The pizza is delicious.}'' When the target is implicitly expressed (e.g., ``\textit{It is overpriced!}''), we can denote the aspect term as a special one named ``\textit{null}''.
    \item \textbf{opinion term} $o$ is the expression given by the opinion holder to express his/her sentiment towards the target. For instance, ``\textit{delicious}'' is the opinion term in the running example ``\textit{The pizza is delicious}''.
    \item \textbf{sentiment polarity} $p$ describes the orientation of the sentiment over an aspect category or an aspect term, which usually belongs to \verb|positive|, \verb|negative|, and \verb|neutral|. 
\end{itemize} 

Note that in the literature, the terminologies of ABSA studies are often used interchangeably, but sometimes they have different meanings according to the context.
For example, ``opinion target'', ``target'', ``aspect'', ``entity'' are usually used to refer to the target on which the opinion is expressed. However, they can be either an aspect category or an aspect term depending on the context. This may cause unnecessary confusion and often makes the literature review incomplete. In this survey, we adopt the most commonly accepted terminologies while also ensuring that similar concepts are clearly distinguishable. 
Therefore, as defined above, we would use ``aspect term'' and ``aspect category'' to differentiate different formats of the aspect, and only use ``target'' or ``aspect'' as a general expression for describing an opinion target. 

\begin{table*}[!t]
    \renewcommand{\arraystretch}{1.3}
    \caption{An overview of different ABSA tasks with different modeling paradigms.} 
    \label{tab:task-paradigm}
    \centering
    \begin{tabular}{c|c|c|c|c|c}
    \hline
    Task & \texttt{SeqClass} & \texttt{TokenClass} & \texttt{MRC} & \texttt{Seq2Seq} & \texttt{Pipeline} \\
    \hline
    Aspect Term Extraction & - & BiLSTM-ATE~\citep{emnlp15-ate-rnn} & QDSL~\citep{aaai21-aope-mrc} & Seq2Seq4ATE~\citep{acl19-ate-seq2label} & -  \\
    Aspect Category Detection & RepLear~\citep{aaai15-acd-zhou} & - & - & - & -  \\
    Opinion Term Extraction & - & RNCRF~\citep{emnlp16-ate-rncrf} & QDSL~\citep{aaai21-aope-mrc} & - & - \\
    Aspect Sentiment Classification & AdaRNN ~\citep{acl14-asc-dong} & - & Dual-MRC~\citep{aaai21-aste-dualmrc} & Gen-ABSA~\citep{acl21-aste-generative-qxp} & - \\
    \hline
    Aspect-Opinion Pair Extraction & - & GTS~\citep{emnlp20-aste-grid} & QDSL~\citep{aaai21-aope-mrc} & GAS~\citep{acl21-gabsa} & - \\
    Aspect Category Sentiment Analysis & AddOneDim~\citep{emnlp18-acsa-joint} & - & - & - &  -\\
    End-to-End ABSA & - & NN-CRF~\citep{emnlp15-uabsa} & Dual-MRC~\citep{aaai21-aste-dualmrc} & Gen-ABSA~\citep{acl21-aste-generative-qxp} & SPAN~\citep{acl19-openabsa} \\
    Aspect Sentiment Triplet Extraction & - & JET~\citep{emnlp20-aste-position} & BMRC~\citep{aaai21-aste-bimrc} & GAS~\citep{acl21-gabsa} & TwoStage~\citep{aaai20-aste} \\
    Aspect-Category-Sentiment Detection & TAS-BERT~\citep{aaai20-tasd} & TAS-BERT~\citep{aaai20-tasd} & - & GAS~\citep{acl21-gabsa} & -\\
    Aspect Sentiment Quad Prediction & - & - & - & Paraphrase~\citep{emnlp21-quad} & Extract-Classify~\citep{acl21-quad-cai} \\
    \hline
    \end{tabular}
\end{table*}

\subsection{ABSA Definition} \label{sec:absa-def}
With the four key sentiment elements defined in the last section, we can give a  definition of ABSA from the perspective of concerned sentiment elements: 
\begin{quote}
    \textbf{Aspect-based sentiment analysis (ABSA)} is the problem to identify sentiment elements of interest for a concerned text item\footnote{In most ABSA benchmark datasets, a sentence is treated as the text item, we thus use ``sentence'' to refer to a concerned text in this paper. However, the reader should be aware that the described methods naturally handle texts with any length.}, either a single sentiment element, or multiple elements with the dependency relation between them.
\end{quote}

We can thus organize various ABSA studies according to the sentiment elements involved.
Depending on whether the desired output is a single sentiment element or multiple coupled elements, we can categorize ABSA tasks into \textbf{single ABSA} tasks and \textbf{compound ABSA} tasks, e.g., aspect term extraction (ATE) is a single ABSA task that aims to extract all aspect terms $a$ given a sentence, while aspect-opinion pair extraction (AOPE) task is a compound ABSA task since it extracts all $(a, o)$ pairs.
From this perspective, we present a new taxonomy for ABSA which systematically organizes existing works from the axes of concerned sentiment elements. We present an overview of different ABSA tasks and representative methods for each task in Fig. \ref{fig:absa-tasks}.

In the light of the above definition, we describe single ABSA tasks in Sec \ref{sec:single-absa} and compound ABSA tasks in Sec \ref{sec:compound-absa}. For each task, we describe what the sentiment elements in the input and output are, what its relation with other tasks is, what the existing solutions are, especially recent progress achieving state-of-the-art performance, as well as general observations and conclusions from previous studies.

\subsection{Modeling Paradigms} \label{sec:modeling-paradigm}
Before describing specific ABSA tasks and their solutions, we introduce several mainstream natural language processing (NLP) modeling paradigms that are commonly employed for ABSA tasks, including Sequence-level Classification (\texttt{SeqClass}), Token-level Classification (\texttt{TokenClass}), Machine Reading Comprehension (\texttt{MRC}), and Sequence-to-Sequence modeling (\texttt{Seq2Seq}).
Each paradigm denotes a general computational framework for handling a specific input and output format. Therefore, by formulating a task as the specific format, the same paradigm can be used to solve multiple tasks \citep{arxiv21-nlp-paradigm}. 
Besides these four unified paradigms which tackle the task in an end-to-end fashion, some complicated ABSA tasks can be solved by the Pipeline (\texttt{Pipeline}) paradigm which pipes multiple models to make the final prediction. In Table \ref{tab:task-paradigm}, we present representative studies of different modeling paradigms for each task.

We denote a dataset corresponding to a certain ABSA task as $\mathcal{D}=\{X_i,Y_i\}_{i=1}^{|\mathcal{D}|}$, where $X_i$ and $Y_i$ are the input and the ground-truth label of the $i$-th data instance\footnote{Since most ABSA tasks are studied in the supervised setting, we mainly describe the modeling paradigm with the supervised learning framework in this section. We also discuss unsupervised settings and methods for specific tasks in later sections.} respectively. We then use such a notation to describe each paradigm.

\subsubsection{Sequence-level Classification (\texttt{SeqClass})}
For the sequence-level classification, a model typically first feeds the input text $X$ into an encoder $\mathbf{Enc}(\cdot)$ to extract the task-specific features, followed by a classifier $\mathbf{CLS}(\cdot)$  to predict the label $Y$: 
\begin{equation}
    Y = \mathbf{CLS}(\mathbf{Enc}(X)), 
\end{equation}
where $Y$ can be represented as one-hot or multi-hot vectors (for single-label and multi-label classification, respectively). 
In the era of deep learning, the encoder $\mathbf{Enc}(\cdot)$ could be convolutional networks~\citep{emnlp14-cnn}, recurrent networks~\citep{nc97-lstm}, or Transformers~\citep{nips17-transformer} for extracting contextual features. 
In some cases, the input text $X$ may contain multiple parts, e.g., for the aspect sentiment classification task, both the sentence and a specific aspect are treated as the input. Then the encoder needs to not only extract useful features, but also capture the interactions between the inputs.
The classifier $\mathbf{CLS}(\cdot)$ is usually implemented as a multi-layer perceptron with a pooling layer to make the classification.

\subsubsection{Token-level Classification (\texttt{TokenClass})}
In contrast to the sequence-level classification that assigns the label to the whole input text, token-level classification (also referred to as sequence labeling or sequence tagging) assigns a label to each token in the input text. 
It also first encodes the input text into contextualized features with an encoder $\mathbf{Enc}(\cdot)$, while employs a decoder $\mathbf{Dec}(\cdot)$ to predict the labels $y_1,...,y_n$ for each token $x_1,...,x_n$ in the input $X$: 
\begin{equation}
    y_1,...,y_n = \mathbf{Dec}(\mathbf{Enc}(x_1,...,x_n)),
\end{equation}
where $\mathbf{Dec}(\cdot)$ can be implemented as either a multi-layer perceptron with a softmax layer, or conditional random fields (CRF)~\citep{icml01-crf}. Different tagging schemes can also be used, e.g., the BIOES tagging scheme (B-beginning, I-inside, O-outside, E-ending, S-singleton) \citep{eacl99-tagging-scheme}.

\subsubsection{Machine Reading Comprehension (\texttt{MRC})}
The \texttt{MRC} paradigm \citep{thesis18-mrc-cdq} extracts continuous text spans from the input text $X$ conditioned on a given query $X_q$. Therefore, ABSA methods with the MRC paradigm need to construct a task-specific query for the corresponding task, i.e., a query denoting what is the desired information. For example, $X_q$ can be constructed as ``\textit{What are the aspect terms?}'' in the ATE task. The original text, as well as the constructed query can then be used as the input to a MRC model to extract the text spans of aspect terms.
It produces the result through predicting the starting position $y_s$ and the ending position $y_e$ of the text span:
\begin{equation}
    y_s, y_e = \mathbf{CLS}(\mathbf{Enc}(X,X_q)),
\end{equation}
where there are typically two linear classifiers, stacked on top of an encoder $\mathbf{Enc}(\cdot)$, for predicting the starting and the ending positions, respectively.

\subsubsection{Sequence-to-Sequence (\texttt{Seq2Seq})}
The sequence-to-sequence (Seq2Seq) framework takes an input sequence $X=\{x_1,...,x_n\}$ as input and aims to generate an output sequence $Y=\{y_1,...,y_m\}$. A classical NLP application with such a paradigm is the machine translation task \citep{nips14-seq2seq}. It is also used for solving ABSA tasks, e.g., directly generating the label sequence or desired sentiment elements given the input sentence. Taking the ATE task as an example, $X$ can be \textit{``The fish dish is fresh''}, and $Y$ can be \textit{``fish dish''} in the natural language form. It typically adopts an encoder-decoder model such as Transformer~\citep{nips17-transformer}:
\begin{equation}
    y_1,...,y_m=\mathbf{Dec}(\mathbf{Enc}(x_1,...,x_n)),
\end{equation}
where the encoder $\mathbf{Enc}(\cdot)$ encodes contextualized features of the input, the decoder $\mathbf{Dec}(\cdot)$ generates a token at each step, based on the encoded input and the previous output.

\begin{table*}[!t]
    \renewcommand{\arraystretch}{1.3}
    \caption{An overview of common ABSA benchmark datasets, listed in chronological order.}
    \label{tab:datasets}
    \centering
    \begin{tabular}{c|c|c|c|c}
    \hline
    Dataset & Language & Major Domains* & Annotations & URL \\
    \hline
    SemEval-2014 \citep{semeval14-absa} & English & \texttt{Lap}, \texttt{Rest} & $a$, $c$, $p$ & \url{https://alt.qcri.org/semeval2014/task4/} \\
    SemEval-2015 \citep{semeval15-absa} & English & \texttt{Lap}, \texttt{Rest} & $a$, $c$, $p$ & \url{https://alt.qcri.org/semeval2015/task12/} \\
    SemEval-2016 \citep{semeval16-absa} & multilingual & \texttt{Elec}, \texttt{Hotel}, \texttt{Rest} & $a$, $c$, $p$  &
    \url{https://alt.qcri.org/semeval2016/task5/} \\
    TOWE \citep{naacl19-towe} & English & \texttt{Lap}, \texttt{Rest} & $a$, $o$ & \url{https://github.com/NJUNLP/TOWE} \\
    ASC-QA \citep{acl19-absa-qa} & Chinese & \texttt{Bag}, \texttt{Cos}, \texttt{Elec} & $a$, $c$, $p$ & \url{https://github.com/jjwangnlp/ASC-QA} \\
    MAMS \citep{emnlp19-absa-mams} & English & \texttt{Rest} & $a$, $c$, $p$ & \url{https://github.com/siat-nlp/MAMS-for-ABSA} \\
    ARTS \citep{emnlp20-asc-tasty} & English & \texttt{Lap}, \texttt{Rest} & $a$, $p$ & \url{https://github.com/zhijing-jin/ARTS_TestSet} \\
    ASTE-Data-V2 \citep{emnlp20-aste-position} & English & \texttt{Lap}, \texttt{Rest}  & $a$, $p$, $o$  &  \url{https://github.com/xuuuluuu/Position-Aware-Tagging-for-ASTE} \\
    ASAP \citep{naacl21-acsc-asap} & Chinese & \texttt{Rest} & $c$, $p$ & \url{https://github.com/Meituan-Dianping/asap} \\
    ACOS \citep{acl21-quad-cai} & English & \texttt{Lap}, \texttt{Rest} & $a$, $c$, $p$, $o$ & \url{https://github.com/NUSTM/ACOS} \\
    ABSA-QUAD \citep{emnlp21-quad} & English & \texttt{Rest}  & $a$, $c$, $p$, $o$  & \url{https://github.com/IsakZhang/ABSA-QUAD} \\
    \hline
    \multicolumn{5}{l}{* domain abbreviations: \texttt{Lap}-laptops, \texttt{Rest}-restaurants, \texttt{Elec}-electronics, \texttt{Cos}-cosmetics}
    \end{tabular}
\end{table*}

\subsubsection{Pipeline Method (\texttt{Pipeline})}
The pipeline method is usually used for tackling compound ABSA tasks due to their complexities. As the name suggests, it sequentially pipes multiple models with possibly different modeling paradigms to obtain the final result. The prediction of the former model is used as the input for the latter model, until the final output is produced. 
For example, the aforementioned AOPE problem aims to extract all (aspect term, opinion term) pairs. Therefore, one straightforward pipeline-style solution is to first use an ATE model to extract potential aspect terms, then employ another model to identify the corresponding opinion terms for each predicted aspect term. The valid predictions can then be organized as (aspect term, opinion term) pairs as the final results.

Compared with unified paradigms described in the previous sections that solve the original problem in an end-to-end manner, the pipeline method is usually easier to implement, since solutions to each sub-problem often already exist. However, it suffers from the error propagation issue, i.e., the errors produced by early models would propagate to the later models and affect the final overall performance. Return to the above example, if the ATE model produces wrong predictions, the final pair extraction results would be incorrect no matter how accurate the second model is. Given the imperfect performance of even simple ABSA tasks, pipeline methods often perform poorly on compound ABSA tasks, especially the complex ones. This often serves as the main motivation to design a unified model to handle the compound ABSA tasks in recent years, as we will discuss in the later sections.

\subsection{Datasets \& Evaluations} \label{sec:dataset}
Annotated datasets play an essential role in the development of ABSA methods. This section presents some commonly used datasets and the corresponding evaluation metrics. An overview of each dataset with its language, data domain, annotated sentiment elements, and URL is summarized in Table \ref{tab:datasets}.

Datasets provided by SemEval-2014~\citep{semeval14-absa}, SemEval-2015~\citep{semeval15-absa}, and SemEval-2016~\citep{semeval16-absa} shared tasks are the most widely used benchmarks in the literature. User reviews from two domains, namely laptops and restaurants, are collected and annotated by the workshop organizers. These datasets contain annotations of aspect categories, aspect terms, and sentiment polarities (although not all of them contain all information), and thus can be directly used for many ABSA tasks such as aspect term extraction or aspect sentiment classification. However, the annotation of opinion terms is lacking, which is later compensated by the dataset provided in \citep{naacl19-towe} for conducting the target-oriented opinion word extraction (TOWE) task. With these annotations available, \citet{emnlp20-aste-position} organize them together (with minor refinements) to obtain the ASTE-Data-V2 dataset, which contains (aspect term, opinion term, sentiment polarity) triplets for each sample sentence\footnote{This dataset is based on the one used in \citet{aaai20-aste}, which they refer to as ASTE-Data-V1. It is further refined with some missing annotations, and the newly obtained dataset is thus called ``V2''.}
. Most recently, two datasets including ACOS~\citep{acl21-quad-cai} and ABSA-QUAD~\citep{emnlp21-quad} for the sentiment quad prediction task have been introduced, where each data instance of them contains the annotations of the four sentiment elements in the quad format.

Besides datasets used for tackling various ABSA tasks, datasets with other special focus have also been introduced. \citet{emnlp19-absa-mams} present a Multi-Aspect Multi-Sentiment (MAMS) dataset where each sentence in MAMS contains at least two aspects with different sentiment polarities, thus making the dataset more challenging. \citet{acl19-absa-qa} propose a Chinese dataset for conducting the ASC task in the QA-style reviews (ASC-QA). \citet{emnlp20-asc-tasty} construct an Aspect Robustness Test Set (ARTS), based on the SemEval-2014 dataset, to probe the robustness of ABSA models. Recently, \citet{naacl21-acsc-asap} release a large-scale Chinese dataset called ASAP (stands for Aspect category Sentiment Analysis and rating Prediction). Each sentence in ASAP is annotated with sentiment polarities of 18 pre-defined aspect categories and the overall rating, so it can also be useful to study the relationship between the coarse-grained and fine-grained sentiment analysis task.

Regarding the evaluation metrics, exact-match evaluation is widely adopted for various tasks and datasets: a prediction is correct if and only if all the predicted elements are the same as the human annotations. Then the typical classification metrics such as accuracy, precision, recall, and F1 scores can be calculated accordingly and used to compare among different methods.

\section{Single ABSA Tasks} \label{sec:single-absa}
We first discuss single ABSA tasks in this section, whose target is to predict a single sentiment element only. As introduced in Sec \ref{sec:four-elements}, there are four tasks (corresponding to four sentiment elements), namely the aspect term extraction (ATE), aspect category detection (ACD), aspect sentiment classification (ASC), and opinion term extraction (OTE). A detailed taxonomy and representative methods are listed in the top branch of Fig. \ref{fig:absa-tasks}.

\subsection{Aspect Term Extraction (ATE)} \label{sec:ate}
Aspect term extraction is a fundamental task of ABSA, aiming to extract explicit aspect expressions on which users express opinions in the given text. For example, two aspect terms ``\textit{pizza}'' and ``\textit{service}'' are supposed to be extracted for the example sentence ``\textit{The pizza is delicious, but the service is terrible.}'' in Table \ref{tab:tasks}. According to the availability of labeled data, ATE methods can be categorized into three types: supervised, semi-supervised, and unsupervised methods.

Supervised ATE problem is often formulated as a token-level classification (i.e., \texttt{TokenClass}) task since the desired aspect terms are usually single words or phrases appeared in the sentence. Therefore, sequence labeling methods based on CRF \citep{ijcai16-ate-wordembed}, RNN \citep{emnlp15-ate-rnn}, and CNN \citep{acl18-ate-decnn} have been proposed. 
Since ATE requires domain-specific knowledge to identify aspects in a given domain, many research efforts are dedicated to improving word representation learning. \citet{ijcai16-ate-wordembed} utilize the dependency path to link words in the embedding space for learning word representations. DE-CNN model proposed by \citet{acl18-ate-decnn} employs a dual embedding mechanism, including general-purpose and domain-specific embeddings. \citet{naacl19-bertabsa} further post-train BERT on domain-specific data to obtain better word representations.
\citet{coling20-ate-pod} design a positional dependency-based word embedding (POD) to consider both dependency relations and positional context.
Specific network designs have been also proposed, e.g., modeling the relation between an aspect and its corresponding opinion expression \citep{emnlp16-ate-rncrf, aaai17-ate-cmla, emnlp17-ate-min, ijcai18-ate-hast}, and transforming the task to a \texttt{Seq2Seq} problem to capture the overall meaning of the entire sentence to predict the aspect with richer contextual information \citep{acl19-ate-seq2label}.

Although achieving satisfactory results (e.g., around 80\% F1 scores on benchmark datasets), supervised ATE methods require a large amount of labeled data, especially when training very sophisticated neural models. This motivates the recent trend on semi-supervised ATE studies. Given a set of labeled ATE data, as well as a (comparatively) large unlabeled dataset (e.g., plain review sentences), data augmentation is an effective solution to produce more pseudo-labeled data for training the ATE model. Various augmentation strategies have been proposed, such as masked sequence-to-sequence generation \citep{acl20-ate-maskedseq2seq}, soft prototype generation \citep{emnlp20-ate-softproto}, and progressive self-training \citep{emnlp21-ate-selftrain}.

Unsupervised ATE task aims to extract aspect terms without any labeled data and has been extensively studied in the literature \citep{widm18-sa-survey}. In the context of neural network based methods, \citet{acl17-ruidan-ate} present an autoencoder model named Attention-based Aspect Extraction (ABAE) that de-emphasizes the irrelevant words to improve the coherence of extracted aspects. Following this direction, \citet{ijcai19-unsupervised-ate} leverage sememes to enhance lexical semantics when constructing the sentence representation. \citet{emnlp19-ate-coupling} propose a neural model to couple both local and global context (LCC+GBC) to discover aspect words.
\citet{acl20-acd-cat} propose a simple solution named CAt where they only use a POS tagger and in-domain word embeddings to extract the aspect terms: the POS tagger first extracts nouns as the candidate aspect, a contrastive attention mechanism is then employed to select aspects. \citet{aaai21-acd} formulate the problem as a self-supervised contrastive learning task to learn better aspect representations.

\begin{table*}[!t]
    \renewcommand{\arraystretch}{1.3}
    \caption{An overview of the input and output for each ABSA task with examples. }
    \label{tab:tasks}
    \centering
    \begin{tabular}{c|c|c|c|c}
    \hline
    Task & Input & Example Input* & Output & Example Output \\
    \hline
    Aspect Term Extraction & $s$ & sentence & $\{a\}$  & \{pizza, service\} \\
    Aspect Category Detection & $s$ & sentence & $\{c\}$  & \{\verb|food|, \verb|service|\} \\
    Aspect Opinion Co-Extraction & $s$ & sentence & $\{a$\}, \{$o\}$ & \{pizza, service\}, \{delicious, terrible\} \\
    \multirow{2}{*}{Target-oriented Opinion Words Extraction} & s, $a_1$ & sentence, pizza & $o_1$  &  delicious \\
    & s, $a_2$ & sentence, service & $o_2$ & terrible \\
    \multirow{2}{*}{Aspect Sentiment Classification} & s, $a_1$ & sentence, pizza & $p_1$  &  \verb|POS| \\
    & s, $a_2$ & sentence, service & $p_2$ & \verb|NEG| \\
    \hline
    Aspect-Opinion Pair Extraction & $s$ & sentence & \{($a$, $o$)\}& (pizza, delicious), (service, terrible) \\
    End-to-End ABSA &  $s$ & sentence & $\{(a, p)\}$ & (pizza, \verb|POS|), (service, \verb|NEG|)  \\
    Aspect Category Sentiment Analysis &  $s$ & sentence & $\{(c, p)\}$ & (\verb|food|, \verb|POS|), (\verb|service|, \verb|NEG|)  \\
    Aspect Sentiment Triplet Extraction & $s$ & sentence & $\{(a, p, o)\}$ & (pizza, \verb|POS|, delicious), (service, \verb|NEG|, terrible)  \\
    Aspect-Category-Sentiment Detection & $s$ & sentence & $\{(c, a, p)\}$ & (\verb|food|, pizza, \verb|POS|), (\verb|service|, service, \verb|NEG|) \\
    \multirow{2}{*}{Aspect Sentiment Quad Prediction} & \multirow{2}{*}{s} & \multirow{2}{*}{sentence} & \multirow{2}{*}{$\{(c, a, p, o)\}$} & (\verb|food|, pizza, \verb|POS|, delicious), \\
    & & & & (\verb|service|, service, \verb|NEG|, terrible) \\
    \hline
    \multicolumn{5}{l}{* We assume the concerned ``\textit{sentence}'' for all example inputs is: ``\textit{The pizza is delicious, but the service is terrible}''.}
    \end{tabular}
\end{table*}

\subsection{Aspect Category Detection (ACD)} \label{sec:acd}
Aspect category detection is to identify the discussed aspect categories for a given sentence, where the categories belong to a pre-defined category set and is often domain-specific~\citep{semeval14-absa}. As shown in Table \ref{tab:tasks}, an ACD method should predict the category \verb|food| and \verb|service| given the sentence. Compared with the ATE task, ACD can be beneficial from two perspectives: Firstly, ATE predicts individual aspect terms while the predicted category of ACD can be regarded as an aggregated prediction, which is more concise to present the opinion target. Secondly, ACD can identify the opinion targets even when they are not explicitly mentioned. For example, given a sentence ``\textit{It is very overpriced and not tasty}'', ACD can detect two aspect categories \texttt{price} and \texttt{food}, whereas ATE is not applicable to such a case.

ACD can be classified into supervised ACD and unsupervised ACD, depending on whether the annotated data is available. The supervised ACD task is usually formulated as a multi-label classification (i.e., \texttt{SeqClass}) problem, treating each aspect category as a label. An early work named RepLearn~\citep{aaai15-acd-zhou} trains the word embedding on a noisy labeled dataset and obtains hybrid features through different feed-forward networks. A logistic regression model is then trained with such features to make the prediction.
Later methods further leverage different characteristics of the task to improve the performance, e.g., using attention mechanism to attend to different parts of the text for different categories \citep{arxiv19-acd-topic}, considering the word-word co-occurrence patterns \citep{tycb18-acd}, and measuring the text matching between the sentence and a set of representative words in each specific category to predict whether a category exists \citep{ecir19-acd-matching}.

To tackle the ACD task in an unsupervised manner, it is often decomposed into two steps: (1) extracting candidate aspect terms, and (2) mapping or clustering the aspect terms to aspect categories in a pre-defined category set \citep{acl17-ruidan-ate}, e.g., clustering ``\textit{pizza}'' and ``\textit{pasta}'' to the aspect category \texttt{food}. 
The first step is essentially the same as tackling the unsupervised ATE problem. For the second step, the most straightforward solution is to manually assign a label for each detected aspect cluster from the first step as the aspect category \citep{acl17-ruidan-ate, ijcai19-unsupervised-ate}, but it is time-consuming and may lead to errors when the detected aspects are noisy. In CAt~\citep{acl20-acd-cat}, the cosine similarity between the sentence vector and the category vector is computed to assign the category label. Most recently, \citet{aaai21-acd} propose a high-resolution selective mapping strategy to improve the mapping accuracy.

\subsection{Opinion Term Extraction (OTE)} \label{sec:ote}
Opinion term extraction (OTE) is the task to identify opinion expressions towards an aspect. Since the opinion term and aspect term always co-occur, solely extracting the opinion term without considering its associated aspect is meaningless. Therefore, depending on whether the aspect term appears in the input or output, OTE can be roughly divided into two tasks: 1) aspect opinion co-extraction (AOCE) and 2) target-oriented opinion words extraction (TOWE).

Aspect opinion co-extraction (AOCE) attempts to predict the aspect and opinion terms together. For the running example in Table \ref{tab:tasks}, the target output of AOCE is thus two aspect terms ``\textit{pizza}'' and ``\textit{service}'', as well as two opinion terms ``\textit{delicious}'' and ``\textit{terrible}''. Note that although two sentiment elements are involved here, AOCE is still a single ABSA task since the dependency relation between the two sentiment elements (e.g., ``\textit{delicious}'' is used to describe ``\textit{pizza}'') is not considered\footnote{In fact, we can directly call this task as ``OTE'' if we neglect the prediction of aspect terms. But purely extracting opinion expressions without their targets is meaningless, leading to the AOCE task.}. Very often, it is formulated as a \texttt{TokenClass} problem with either two label sets to extract aspect and opinion terms separately \citep{taslp19-aoce-yu}, or a unified label set (e.g., \{B-A, I-A, B-O, I-O, N\} denotes the beginning (B) or inside (I) of an aspect (-A) or opinion term (-O), or none of them (N)) to extract both sentiment elements simultaneously \citep{acl18-ate-recursive, emnlp20-aoce-wwy}. Considering the close relationship between the aspect and opinion, the main research question of AOCE is how to model such a dependency relation. Various models have been developed to capture the aspect-opinion dependency, including dependency-tree based models \citep{ijcai16-ate-wordembed, emnlp16-ate-rncrf}, attention-based models \citep{aaai17-ate-cmla, emnlp17-ate-min, ijcai18-ate-hast}, and the models considering the syntactic structures to explicitly constrain the prediction \citep{taslp19-aoce-yu, emnlp20-aoce-wwy}.

On the other hand, TOWE aims to extract the corresponding opinion terms given a specific aspect term within the text \citep{naacl19-towe}. As shown in Table \ref{tab:tasks}, an aspect of interest (e.g., ``pizza'') is assumed to be given with the sentence, then a TOWE model aims to predict the corresponding opinion term (e.g., ``delicious'').
TOWE is also often formulated as a \texttt{TokenClass} problem towards the input sentence, 
whereas the main research problem becomes how to model the aspect-specific representation in the input sentence to extract the corresponding opinions. \citet{naacl19-towe} propose a neural model to incorporate the aspect information via an inward-outward
LSTM to generate the aspect-fused context.
Later methods manage to enhance the accuracy of extraction from several aspects: \citet{aaai20-towe-transfer} utilize the general sentiment analysis dataset to transfer the latent opinion knowledge for tackling TOWE. \citet{emnlp20-towe-syntax} leverage the syntactic structures such as the dependency tree-based distance to the aspect to help identify the opinion terms. \citet{emnlp21-towe-position} empirically evaluate the importance of positional embeddings based on various text encoders and find that BiLSTM-based methods have an inductive bias appropriate for the TOWE task, and using a GCN \citep{iclr17-gcn} to explicitly consider the structure information only brings minor performance gains.

\subsection{Aspect Sentiment Classification (ASC)} \label{sec:asc}
Aspect sentiment classification (ASC), also called aspect-based/-targeted/-level sentiment classification, aims to predict the sentiment polarity for a specific aspect within a sentence. Generally, the aspect can be instantiated as either aspect term or aspect category, yielding two ASC problems: aspect term-based sentiment classification and aspect category-based sentiment classification. Regardless of some subtle differences (e.g., the given aspect term comes from the sentence, then its position information can be exploited), 
the main research question underlying these two settings is the same: how to appropriately exploit the connection between the aspect (term/category) and the sentence context to classify the sentiment. In fact, some works consider these two subtasks at the same time and tackle them seamlessly with the same model \citep{emnlp16-asc-atae, aaai18-asc-yitay, acl18-asc-cnn, ijcai19-asc-early}. Therefore, we do not specifically differentiate these two subtasks in this section and use ``aspect'' to refer to either an aspect term or an aspect category. 

Deep learning based ASC has attracted a lot of interest  and a variety of neural network based models have been proposed and brought large performance improvements \citep{ijcai15-asc-nn, access19-absa-survey, tcss20-absa-survey}.
To model the interaction between the aspect and sentence context, pioneering neural models such as TC-LSTM ~\citep{coling16-asc-effective-lstm} employ relatively simple strategies such as concatenation to fuse the aspect information with the sentence context. Based on the intuition that different parts of the sentence play different roles for a specific aspect, the attention mechanism is widely employed to obtain aspect-specific representations \citep{emnlp16-asc-atae, aaai18-asc-yitay, eacl17-asc-attention, cikm17-asc-heat, coling18-asc-ruidan, emnlp20-asc-xulu}. A representative work is Attention-based LSTM with Aspect Embedding (ATAE-LSTM) model proposed by \citet{emnlp16-asc-atae} which appends the aspect embedding to each word vector of the input sentence for computing the attention weight, and an aspect-specific sentence embedding can be computed accordingly to classify the sentiment. The following methods design more complicated attention mechanisms to learn better aspect-specific representations, for instance, IAN~\citep{ijcai17-asc-ian} interactively learns attention in the aspect and sentence, and generate the representations for them separately. 
Apart from the LSTM network, other network structures have also been explored for supporting the attention mechanism, including the CNN-based network \citep{acl18-asc-cnn, acl18-asc-lx-tnet}, memory network \citep{emnlp16-asc-memory, emnlp17-asc-ram}, and the gated network \citep{aaai16-asc-gated, acl18-asc-cnn}. 
Recently, pre-trained language models have become the mainstream building block for the ASC task \citep{naacl19-utilizing, naacl19-bertabsa, aaai21-asc-context-bert, naacl21-asc-roberta}. For example, \citet{naacl19-utilizing} transform the ASC task as a sentence pair classification problem by constructing an auxiliary sentence, which can better utilize the sentence pair modeling ability of BERT.

Another line of the ASC research explicitly models the syntactic structure of the sentence to make the prediction, since the structural relation between the aspect and its associated opinion often indicates the sentiment orientation. In fact, earlier machine learning based ASC systems already take mined syntactic trees as features for the classification \citep{semeval14-xrce, semeval14-nrc}. However, as the dependency parsing itself is a challenging NLP task, ASC methods with inaccurate parsers did not show clear advantages than other methods \citep{naacl21-asc-roberta}. Thanks to the improvements from the neural network based dependency parsing in recent years, more accurate parse trees have brought significant improvements for the dependency-based ASC model. For example, \citet{emnlp19-asc-cdt} and \citet{emnlp19-asgcn} employ the graph neural network (GNN) \citep{aiopen20-gnn-survey} to model the dependency tree for exploiting the syntactical information and word dependencies. Following this direction, a variety of GNN-based methods have been proposed to explicitly leverage the syntactic information \citep{emnlp19-syntax-gat, acl20-asc-dgedt, emnlp20-bigcn, acl20-relational, emnlp20-latent-graph, acl21-asc-dualgcn, naacl21-asc-graphmerge, naacl21-tgcn, emnlp21-asc-aclt}. 
Besides the syntactic structure inside the sentence, other structural information has also been considered. \citet{emnlp16-acsc-ruder} model the relation between multiple review sentences, with the assumption that they build and elaborate upon each other and thus their sentiments are also related. 
Similarly, \citet{acl20-asc-document} consider the document-level sentiment preference to fully utilize the information in the existing data to improve the ASC performance.

\section{Compound ABSA Tasks} \label{sec:compound-absa}
\begin{figure}[!t]
\centering
\includegraphics[width=\linewidth]{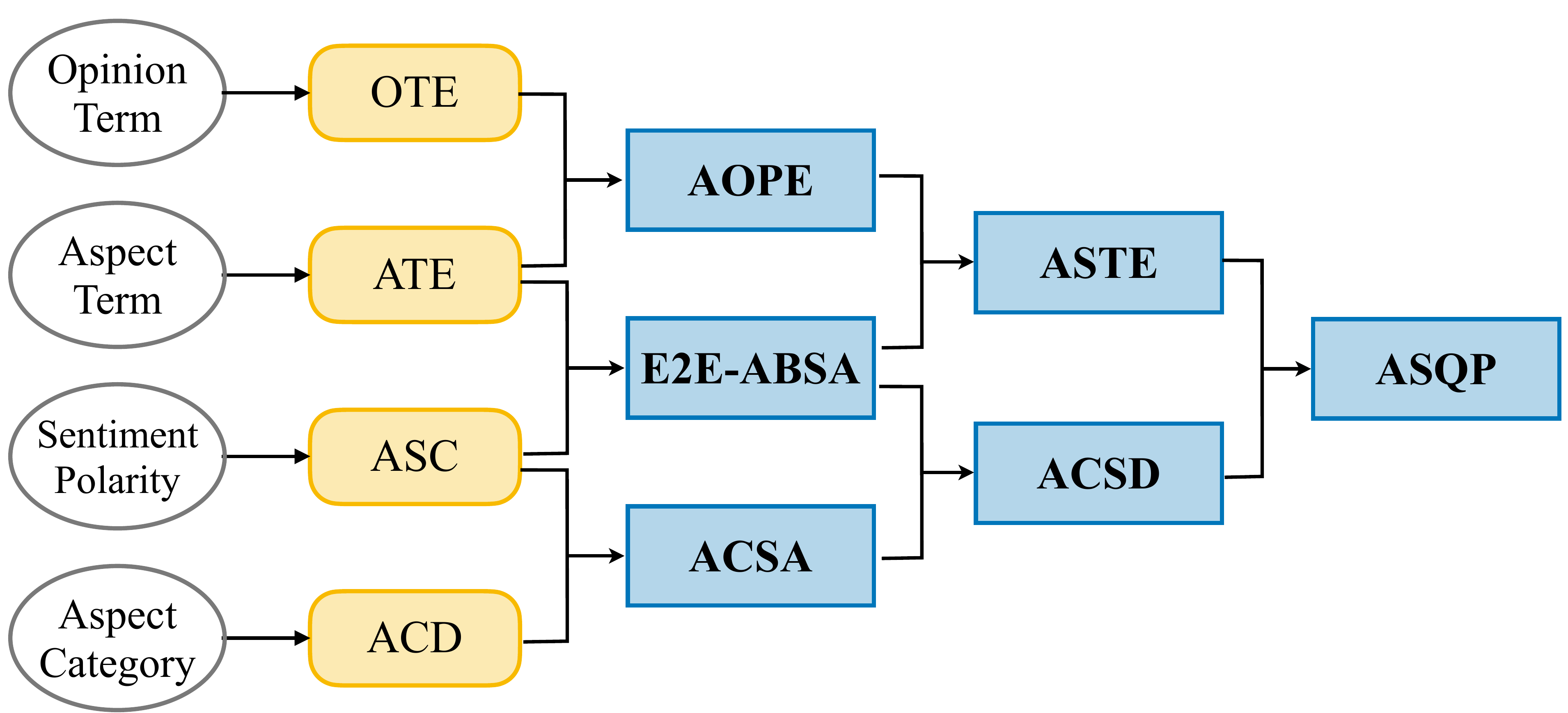}
\caption{The relations between the four sentiment elements, single ABSA tasks, and compound ABSA tasks.} 
\label{fig:compound-taxonomy}
\end{figure}

We then describe compound ABSA tasks whose target involves multiple sentiment elements. A detailed task taxonomy and representative methods are shown in the bottom branch of Fig. \ref{fig:absa-tasks}. Very often, these tasks can be treated as integrated tasks of the aforementioned single ABSA tasks. However, the goal of these compound tasks is not only the extraction of multiple sentiment elements, but also coupling them by predicting the elements in the pair (i.e., two elements), triplet (i.e., three elements), or even quad (i.e., four elements) format. Fig.~\ref{fig:compound-taxonomy} shows the relation between these tasks.
Considering the inter-related dependency of the four sentiment elements, providing an integrated solution is a promising direction. Many research efforts have been made recently, which we systematically review in this section.

\subsection{Aspect-Opinion Pair Extraction (AOPE)} \label{sec:AOPE}

\begin{table*}[!t]
    \renewcommand{\arraystretch}{1.3}
    \caption{Demonstration of the joint and unified tagging methods for the E2E-ABSA task.} 
    \label{tab:e2e-absa}
    \centering
    \begin{tabular}{ccccccccccccc}
    \hline
     & The & AMD & Turin & Processor & seems & to & always & perform & better & than & Intel & . \\
    \hline
    \multirow{2}{*}{\textbf{Joint}} & O & B & I & E & O & O & O & O & O & O & S & O \\   
    & O & POS & POS & POS & O & O & O & O & O & O & NEG & O  \\
    \hline
    \textbf{Unified} & O & B-POS & I-POS & E-POS & O & O & O & O & O & O & S-NEG & O \\
    \hline
    \end{tabular}
\end{table*}

As discussed in Sec \ref{sec:ote}, studies of the aspect opinion co-extraction (AOCE) task often found that the extraction of each element can mutually reinforce each other. However, the output of the AOCE task contains two separate sets: an aspect set and an opinion set. The corresponding pairwise relation is neglected. This motivates the task of aspect-opinion pair extraction (AOPE), aiming to extract the aspect and opinion terms in pairs so as to provide a clear picture of what the opinion target is and what the corresponding opinion expression is \citep{acl20-aope-spanmlt, acl20-aope-sync}. 

To tackle AOPE, one can adopt the \texttt{pipeline} approach to decouple it into several subtasks and pipe them together to obtain the aspect-opinion pairs. One solution is to first conduct the AOCE task for obtaining the aspect and opinion sets, then employ a classification model to pair the potential aspect and opinion terms, i.e., classify whether an aspect-opinion pair is valid. An alternative method is to first extract the aspect (i.e., the ATE task), then identify the corresponding opinion terms for each predicted aspect term (i.e., the TOWE task). \citet{aaai21-aope-mrc} take the second approach with the \texttt{MRC} paradigm where they first use an MRC model to extract all aspect terms, then for each extracted aspect term, a query is constructed for another MRC model to identify the text span of the corresponding opinion term.

Efforts have also been made to tackle AOPE in a unified manner, for alleviating the potential error propagation of the pipeline approach. \citet{emnlp20-aste-grid} propose a grid tagging scheme (GTS): for each word pair, the model predicts whether they belong to the same aspect, the same opinion, the aspect-opinion pair, or none of the above. Then the original pair extraction task is transformed into a unified \texttt{TokenClass} problem.  \citet{acl20-aope-spanmlt} treat the problem as a joint term and relation extraction, and design a span-based multi-task learning (SpanMlt) framework to jointly extract the aspect/opinion terms and the pair relation: a span generator is first used to enumerate all possible spans, then two output scorers assign the term label and evaluate the pairwise relations. Similarly, \citet{acl20-aope-sync} propose a model containing two channels to extract aspect/opinion terms, and the relations respectively. Two synchronization mechanisms are further designed to enable the information
interaction between two channels. More recently, syntactic and linguistic knowledge is also considered for improving the extraction performance \citep{ijcai21-aope-syntax}.

\subsection{End-to-End ABSA (E2E-ABSA)} \label{sec:e2e-absa}
Given a sentence, End-to-End ABSA is the task of extracting the aspect term and its corresponding sentiment polarity simultaneously, i.e., extracting the $(a, p)$ pairs\footnote{Since extracting mentioned aspects and classifying their sentiments lies in the core of ABSA problem~\citep{hlt12-liu-bing-sa}, it is often directly referred to as the ``ABSA problem''. In recent years, to differentiate this task from the general ABSA problem (consisting of multiple tasks), it is called end-to-end ABSA~\citep{wnut19-exploiting, acl19-uabsa-imn} or unified ABSA~\citep{acl20-uabsa-racl, acl21-gabsa}. Following this convention, we thus take the name E2E-ABSA here to denote this task.}.
It can be naturally broken down into two sub-tasks, namely ATE and ASC \citep{acl19-openabsa}, and an intuitive \texttt{pipeline} method is to conduct them sequentially.
However, detecting the aspect boundary and classifying the sentiment polarity can often reinforce each other. Taking the sentence ``\textit{I like pizza}'' as an example, the context information ``\textit{like}'' indicates a positive sentiment and also implies that the following word ``\textit{pizza}'' is the opinion target. Inspired by such an observation, many methods have been proposed to tackle the problem in an end-to-end manner.

These end-to-end methods can be generally divided into two types~\citep{emnlp13-uabsa, emnlp15-uabsa}, as shown in Table \ref{tab:e2e-absa}. The first ``\textbf{joint}'' method exploits the relation between two subtasks via training them jointly within a multi-task learning framework \citep{acl19-uabsa-doer, acl19-uabsa-imn, acl20-uabsa-racl, emnlp20-uabsa-grace, emnlp21-uabsa-iterative}. Two label sets including the aspect boundary label (the first row) and the sentiment label (the second row) are adopted to predict the two types of sentiment elements. Then the final prediction is derived from the combination of the outputs of two subtasks.
Another type of method dismisses the boundary of these two subtasks and employs a ``\textbf{unified}'' (also called collapsed) tagging scheme to denote both sentiment elements in the tag of each token \citep{ijcnn18-uabsa, aaai19-uabsa-lx, wnut19-exploiting}. As shown in the last row of Table \ref{tab:e2e-absa}, the tag for each token now contains two parts of information: the first part \{B, I, E, S, O\} denotes the boundary of the aspect (B-beginning, I-inside, O-outside, E-ending, S-singleton) \citep{eacl99-tagging-scheme}, the second part \{POS, NEG, NEU\} is the sentiment polarity of the corresponding token. For instance, B-NEG refers to the beginning of an aspect whose sentiment is negative. By using a collapsed label scheme, the E2E-ABSA task can be tackled with the \texttt{TokenClass} paradigm via a standard sequence tagger~\citep{wnut19-exploiting}.

Whichever type of method is adopted, some ideas are often shared and appear frequently in different models. For example, considering the relation between the aspect boundary and sentiment polarity has shown to be an important factor~\citep{aaai19-uabsa-lx}. As opinion terms provide indicative clues for the appearance of aspect terms and the orientation of the sentiment, opinion term extraction is often treated as an auxiliary task \citep{aaai19-uabsa-lx, acl19-uabsa-imn, emnlp21-uabsa-iterative, emnlp21-uabsa-self}. For example, the relation-aware collaborative learning (RACL) framework \citep{acl20-uabsa-racl} explicitly models the interactive relation of three tasks with a relation propagation mechanism to coordinate these tasks. \citet{emnlp21-uabsa-iterative} further design a routing algorithm to improve the knowledge transfer between these tasks. Document-level sentiment information is also used to equip the model with coarse-grained sentiment knowledge, so as to better classify the sentiment polarity \citep{acl19-uabsa-imn, emnlp21-uabsa-iterative}.

Regarding these three types of methods for tackling E2E-ABSA (i.e., pipeline, joint, and unified method), it is still unclear which one is the most suitable. Early works such as \citep{emnlp13-uabsa} found that the pipeline method performs better, but \citet{aaai19-uabsa-lx} show that using a tailor-made neural model with the unified tagging scheme gives the best performance. Later, \citet{wnut19-exploiting} further verify that using a simple linear layer stacked on top of the pre-trained BERT model with the unified tagging scheme can achieve promising results, without complicated model design. More recently, research works based on either pipeline \citep{aaai21-aste-dualmrc, aaai21-aste-bimrc}, unified \citep{emnlp21-findings-uabsa-qa}, or the joint method \citep{emnlp21-uabsa-iterative} all achieve good performance, i.e., around 70\% F1 scores on benchmark datasets. This makes the comparison between different types of methods unclear and needs further exploration.

\begin{figure*}[!t]
\centering
\includegraphics[width=0.95\linewidth]{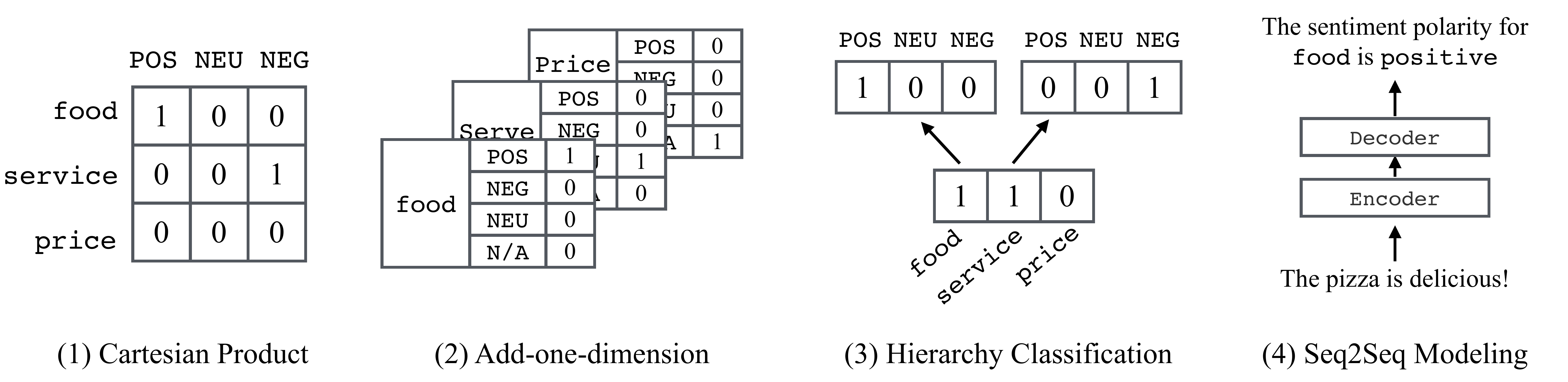}
\caption{Demonstrations of the four types of unified methods for the ACSA task.}
\label{fig:acsa-types}
\end{figure*}

\subsection{Aspect Category Sentiment Analysis (ACSA)} \label{sec:acsa}
Aspect category sentiment analysis (ACSA) aims to jointly detect the discussed aspect categories and their corresponding sentiment polarities. For example, an ACSA model is expected to predict two category-sentiment pairs (\texttt{food}, \texttt{POS}) and (\texttt{service}, \texttt{NEG}) for the example in Table~\ref{tab:tasks}. Though ACSA is similar to the E2E-ABSA task (only the format of the aspect is different), the results of ACSA can be provided regardless of whether the aspect is implicit or explicitly mentioned in the sentence, thus ACSA is widely used in industries \citep{naacl21-acsc-asap}.

The most straightforward method to tackle ACSA is the \texttt{pipeline} approach: first detecting the mentioned aspect categories (i.e., the ACD task), then predicting the sentiment polarities for those detected categories (i.e., the ASC task). However, the detection of a subset of the aspect categories appearing in the sentence is nontrivial, as discussed in Sec \ref{sec:acd}. The errors from the first step would severely limit the performance of the overall pair prediction. Moreover, the relations between these two steps are ignored, which is found to be important for both tasks \citep{emnlp19-can}. In fact, performing these two tasks in the multi-task learning framework has shown to be beneficial for each separate task \citep{aaai18-sentilstm, emnlp19-can, emnlp20-acsa-incremental}.

In essence, the ACD task is a multi-label classification problem (treating each category as a label), and the ASC task is a multi-class classification problem (where each sentiment polarity is a class) for each detected aspect category. As shown in Fig.~\ref{fig:acsa-types}, existing methods of tackling ACSA in a unified manner can be roughly categorized into four types: (1) Cartesian product, (2) add-one-dimension, (3) hierarchy classification, and (4) Seq2Seq modeling.
The Cartesian product method enumerates all possible combinations of category-sentiment pairs by a Cartesian product. Then a classifier takes both the sentence and a specific category-sentiment pair as input, the prediction is thus a binary value, indicating whether such a pair holds \citep{aaai20-tasd}. 
However, it generates the training set several times larger than the original one, greatly increasing the computation cost. An alternative solution is to add one extra dimension to the prediction of the aspect category. Previously, for each aspect category, we predict its sentiment polarity, which normally has three possibilities: positive, negative, and neutral. \citet{emnlp18-acsa-joint} add one more dimension called ``N/A'' denoting whether the category appears in the sentence or not, thus handling the ACSA in a unified way.  

\citet{coling20-acsa-hiergcn} propose a hierarchical GCN-based method named Hier-GCN: a lower-level GCN first captures the relations between categories, then a higher-level GCN is used to capture the relations between categories and category-oriented sentiments. Finally, an integration module takes the interactive features as inputs to perform the hierarchy prediction.
Similarly, \citet{ccl20-acsa} utilize a shared sentiment prediction layer to share the sentiment knowledge between different aspect categories to alleviate the data deficiency issue.
Recently, \citet{emnlp21-acsa-seq2seq} adopts the \texttt{Seq2Seq} modeling paradigm to tackle the ACSA problem. Based on a pre-trained generative model, they use natural language sentences to represent the desired output (see Fig.~\ref{fig:acsa-types}(4)) which outperforms previous classification type models. Moreover, the experimental results suggest that such paradigm can better utilize the pre-trained knowledge and have large advantages in few-shot and zero-shot settings.

\subsection{Aspect Sentiment Triplet Extraction (ASTE)}
\begin{figure*}[!t]
\centering
\includegraphics[width=\linewidth]{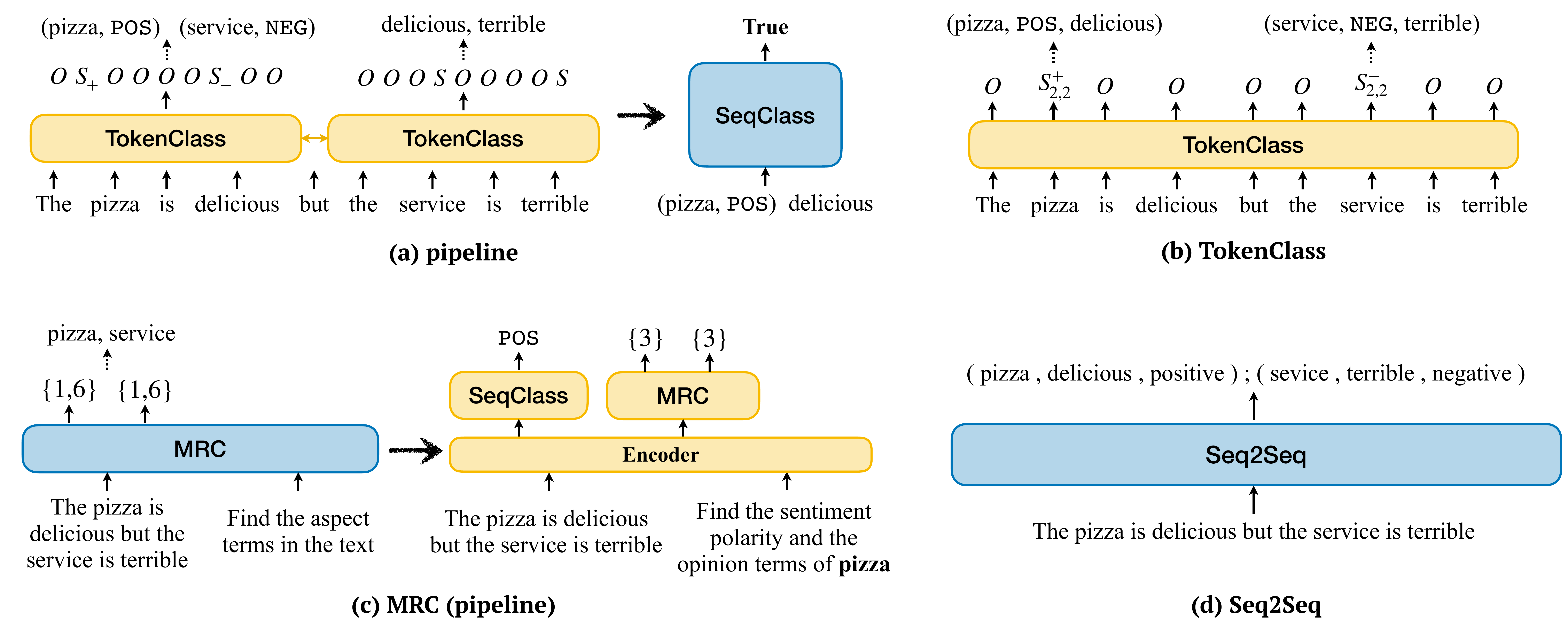}
\caption{Different modeling paradigms for tackling the ASTE task, where (a), (b), (c), and (d) are simplified illustrations of the methods proposed in TwoStage~\citep{aaai20-aste}, JET~\citep{emnlp20-aste-position}, BMRC~\citep{aaai21-aste-bimrc}, and GAS~\citep{acl21-gabsa} respectively.}
\label{fig:aste-methods}
\end{figure*}

The aspect sentiment triplet extraction (ASTE) task attempts to extract $(a, o, p)$ triplets from the given sentence, which tells what the opinion target is, how its sentiment orientation is, and why such a sentiment is expressed (through the opinion term) \citep{aaai20-aste}. Therefore, a model which can predict opinion triplets shows more complete sentiment information, compared with previous models working for individual tasks. The ASTE task has attracted lots of attention in recent years. A variety of frameworks with different paradigms have been proposed for the ASTE task, we show some representative works of each paradigm in Fig.~\ref{fig:aste-methods}.

\citet{aaai20-aste} first introduce the ASTE task and propose a two-stage \texttt{pipeline} method for extracting the triplets. As shown in Fig.~\ref{fig:aste-methods}(a), two sequence tagging models are first performed to extract aspects with their sentiments, and the opinion terms respectively. In the second stage, a classifier is utilized to find the valid aspect-opinion pairs from the predicted aspects and opinions and finally construct the triplet prediction. To better exploit the relations of multiple sentiment elements, many unified methods have been proposed. \citet{emnlp20-aste-mtl} present a multi-task learning framework including aspect term extraction, opinion term extraction, and sentiment dependency parsing tasks. Then heuristic rules are applied to produce the sentiment triplets from the predictions of those subtasks. 
Another potential direction is to design unified tagging schemes to extract the triplet in one-shot \citep{emnlp20-aste-position, emnlp20-aste-grid}: JET model proposed by \citet{emnlp20-aste-position} utilizes a position-aware tagging scheme which extends the previous unified tagging scheme of the E2E-ABSA task \citep{aaai19-uabsa-lx} with the position information of the opinion term, as depicted in Fig.~\ref{fig:aste-methods}(b). Similarly, \citet{emnlp20-aste-grid} extend the grid tagging scheme (GTS) for the AOPE task described in Sec \ref{sec:AOPE} to also make a prediction on the sentiment polarity. 
Since those methods rely on the interactions between word pairs, they may not perform well when the aspect terms or the opinion terms are multi-word expressions. Motivated by this observation, \citet{acl21-aste-span} propose a span-level interaction model which explicitly considers the interactions between the whole spans of aspects and those of opinions to improve the performance.

Other modeling paradigms such as \texttt{MRC} (see Fig.~\ref{fig:aste-methods}(c)) \citep{aaai21-aste-bimrc, aaai21-aste-dualmrc} and \texttt{Seq2Seq} modeling (see Fig.~\ref{fig:aste-methods}(d)) \citep{acl21-aste-generative-qxp, acl21-gabsa, emnlp21-aste-pointer, tnnls21-aste-nonautoregressive} have also been employed for tackling ASTE. \citet{aaai21-aste-dualmrc} transform the original problem as two MRC tasks by designing specific queries: the first MRC model is used to extract the aspect terms, the second MRC model then predicts the corresponding opinion term and sentiment polarity. \citet{aaai21-aste-bimrc} take a similar approach while they use a bidirectional MRC framework: one predicts aspect term then opinion term, another first predicts the opinion then the aspect. 
\texttt{Seq2Seq} modeling provides an elegant solution to make the triplet prediction in one shot. \citet{acl21-gabsa} transform the original task as a text generation problem and propose two modeling paradigms including the annotation style and extraction style for predicting the sentiment triplets. \citet{acl21-aste-generative-qxp} and \citet{emnlp21-absa-spdaug} take the sentence as input and treat the pointer indices as the target. Then to predict the aspect term (or the opinion term), the target becomes predicting the starting index and ending index of the term. \citet{tnnls21-aste-nonautoregressive} presents a non-autoregressive decoding (NAG-ASTE) method which models the ASTE task as an unordered triplet set prediction problem.

\subsection{Aspect-Category-Sentiment Detection (ACSD)} \label{sec:acsd}
Although the aspect category and aspect term can both serve as the opinion target when analyzing aspect-level sentiment, the sentiment often depends on both of them \citep{wassa18-tasd-wild}. To capture such a dual dependence, \citet{aaai20-tasd} propose to detect all (aspect category, aspect term, sentiment polarity) triplets for a given sentence\footnote{The authors call ``aspect category'' and ``aspect term'' as ``aspect'' and ``target'' respectively in the original paper. Here we use unified terminologies to refer to those sentiment elements.}. They separate the joint prediction task into two subtasks on the basis of (aspect category, sentiment polarity) pairs, whose idea is similar to the ``Cartesian Product'' for the ACSA task described in Sec \ref{sec:acsa}. Therefore, given a sentence with a specific combination of the aspect category and sentiment, the remaining problems are: whether any aspect terms exist for such a combination, and if so, what the aspect terms are? The former one can be formulated as a binary \texttt{SeqClass} task, and the latter becomes a conditional \texttt{TokenClass} problem. For example, given the sentence ``\textit{The pizza is delicious}'' with (\verb|food|, \verb|POS|) pair, the first subtask would predict that this combination exists and the sequence labeling model should extract ``\textit{pizza}'' as the corresponding aspect term. Then a triplet (\verb|food|, \verb|POS|, \textit{pizza}) can be output as a prediction. However, when receiving the same sentence with the (\verb|service|, \verb|POS|) pair as input, the first subtask is supposed to predict this combination does not exist. 
The overall training objective can be the combined loss of these two subtasks.

Following this direction, \citet{kbs21-tasd-mejd} propose a model called MEJD which handles the task by using the sentence and a specific aspect category as input, then the remaining problems becomes: (1) predicting the sentiment polarity for the given category (i.e., a \texttt{SeqClass} problem), and (2) extract the corresponding aspect terms if exist (i.e., a \texttt{TokenClass} problem). Since a specific aspect category may not always exist in the concerned sentence, MEJD adds an extra dimension ``N/A'' in the \texttt{SeqClass} task, sharing the similar idea of the ``add-one-dimension'' method \citep{emnlp18-acsa-joint} introduced in Sec \ref{sec:acsa}. Therefore, when the classification model outputs ``N/A'', it shows that there is no triplet related to the category in the input. Moreover, a GCN with an attention mechanism is employed in MEJD to capture the dependency between the aspect and the context.

As the number of predefined aspect categories for a specific domain is generally small, the aforementioned methods can decompose the original ACSD task by combining the sentence with each category as the input. Instead, \citet{acl21-gabsa} tackle the problem in a \texttt{Seq2Seq} manner where they append the desired sentiment elements in the original sentence and treat it as the target sequence for a generation model to learn the mapping relation. \citet{emnlp21-quad} further design a Paraphrase model which constructs a natural language sentence containing all the sentiment elements as the target sequence for the sequence-to-sequence learning.

\subsection{Aspect Sentiment Quad Prediction (ASQP)}
The primary motivation of various compound ABSA tasks discussed above is to capture more detailed aspect-level sentiment information, either in the format of pair extraction (e.g., AOPE) or triplet extraction (e.g., ASTE). Although they can be useful under different scenarios, a model which can predict the four sentiment elements in one shot is supposed to provide the most complete aspect-level sentiment structure. This leads to the aspect sentiment quad prediction (ASQP) task proposed recently\footnote{It is also called Aspect-Category-Opinion-Sentiment (ACOS) quadruple extraction task in \citep{acl21-quad-cai}.} \citep{acl21-quad-cai, emnlp21-quad}, aiming to predict all the four sentiment elements in the quadruplet form given a text item. Returning to the example in Table~\ref{tab:tasks}, two sentiment quads are expected: (\verb|food|, pizza, \verb|POS|, delicious) and (\verb|service|, service, \verb|NEG|, terrible).

\citet{acl21-quad-cai} study the ASQP task with an emphasis on the implicit aspects or opinions. The authors argue that implicit aspects or opinions appear frequently in real-world scenarios, and use ``null'' to denote them in the sentiment quads. They introduce two new datasets with sentiment quad annotations and construct a series of \texttt{Pipeline} baselines by combining existing models to benchmark the task.
\citet{emnlp21-quad} propose a Paraphrase modeling strategy to predict the sentiment quads in an end-to-end manner. By combining the annotated sentiment elements with a pre-built template and using the obtained natural language sentence as the target sequence, they transform the original quad prediction task to a text generation problem and tackle it via the \texttt{Seq2Seq} modeling paradigm. Therefore, the label semantics (i.e., the meaning of the sentiment elements) can be fully exploited. Following this direction, later methods further formulate the task as generating opinion trees \citep{acl22-seq2path, ijcai22-opinion-tree} or structured schema \citep{acl22-uie}.

Compared to other ABSA tasks, ASQP is the most complete and also the most challenging task. The main difficulty lies in the accurate coupling of different sentiment elements. Given the importance of the it and the potential large improvement space (e.g., the current best-performing models only achieve about 40\% F1 scores on benchmark datasets), we expect to see more related studies in the future.

\section{ABSA with Pre-trained Language Models} \label{sec:plm-absa}

Conventional neural ABSA models usually couple the pre-trained word embeddings, such as Word2Vec~\citep{nips13-word2vec} and GloVe~\citep{emnlp14-glove}, with a well-designed task-specific neural architecture. Despite their effectiveness compared with early feature-based models, the improvement from these models gradually reached a bottleneck. One reason is that the context-independent word embeddings are insufficient for capturing the complex sentiment dependencies in the sentence. Besides, the sizes of existing ABSA datasets do not support the training of very complicated architecture. In recent years, pre-trained language models (PLMs) such as BERT~\citep{naacl19-bert} and RoBERTa~\citep{arxiv19-roberta} have brought substantial improvements on a wide range of NLP tasks. Naturally, they are also introduced for further improving the performance of the ABSA problem. 

The initial works~\citep{naacl19-bertabsa,arxiv19-attentional,lrec20-adapt} do not spend too much effort on task-specific model designs, but simply introduce the contextualized embeddings from PLMs as the replacement of word embeddings. Given the rich knowledge learned in the pre-training stage, simply utilizing such contextualized embeddings already brings in a large performance gain. For instance, \citet{wnut19-exploiting} investigate the usage of stacking several standard prediction layers on top of a PLM for the E2E-ABSA task. They find that using the simplest linear classification layer with a PLM can outperform previous carefully-designed neural ABSA models. Similarly, simply concatenating the given sentence and a concerned aspect as the input to PLMs and utilizing the sentence-level output (e.g., representations corresponding to the [CLS] token for BERT) establishes new state-of-the-art results of the ASC task \citep{naacl19-bertabsa}. Moreover, the authors show that further post-training the model on the domain and task related data can capture better domain- and task-specific knowledge, thus leading to better performance. 

However, simply adopting PLMs as the context-aware embedding layer might be insufficient. From the perspective of ABSA tasks, complicated tasks often require not only the identification of the sequence- or token-level labels but also the dependency relations between them, it thus needs more designs to make full use of the contextualized embeddings from PLMs. From the perspective of PLMs, the rich knowledge learned in the pre-training stage might not be sufficiently induced and utilized for the concerned ABSA tasks.
To this end, many efforts have been made on better adapting the PLMs for different downstream ABSA tasks. An early attempt is \citet{naacl19-utilizing} where they transform the ASC as a sentence pair classification task. Motivated by the observation that BERT has an advantage in tackling sentence pair classification problems such as question answering, they construct an auxiliary sentence for each aspect and feed the original sentence and the constructed sentence to BERT, achieving much better performance than previous works. Following similar intuition, \citet{aaai21-aope-mrc,aaai21-aste-bimrc,aaai21-aste-dualmrc} solve the AOPE task and ASTE task via the \texttt{MRC} modeling paradigm. By decomposing the original task as a series of MRC processes, the pairwise relation is thus naturally captured via the query-answer matching.
Another line of work is to utilize the pre-trained generative models such as BART~\citep{acl20-bart} and T5~\citep{jmlr20-t5} to solve various ABSA tasks \citep{acl21-gabsa, acl21-aste-generative-qxp,emnlp21-quad, emnlp21-acsa-seq2seq}. By transforming the original task to a \texttt{Seq2Seq} problem, the label semantics (i.e., the meaning of the desired sentiment elements) can be appropriately incorporated.

In addition to serving as the backbone of ABSA models, PLMs can benefit tackling ABSA tasks from other aspects. For example, the language modeling task used in the pre-training stage of PLMs often brings in the capability of performing generative data augmentation. \citet{acl20-ate-maskedseq2seq} employ PLMs as a conditional text generator and design a mask-then-predict strategy to generate the augmented training sentences for the ATE task. \citet{emnlp21-absa-spdaug} do not borrow the external linguistic resources but utilize the PLMs to achieve semantic-preserved augmentation in a generative manner, obtaining clear improvements over the baseline method on a series of ABSA tasks. 
Another interesting but largely ignored role of PLMs is to provide better dependency trees for various ABSA models, e.g., methods discussed in Sec~\ref{sec:asc}. Explicitly utilizing the semantic relation can be beneficial for many ABSA tasks, but their performance heavily depends on the accuracy of the adopted dependency tree \citep{semeval14-xrce, semeval14-nrc, naacl21-asc-roberta}. As the first attempt, \citet{acl20-asc-perturbed} discover the dependency parse tree from PLMs with a tailor-made probing method and feed the obtained tree into a dependency-based ABSA model, achieving better ASC results than the model using the tree from the off-the-shelf parsers. Following them, \citet{naacl21-asc-roberta} fine-tune the PLMs with the ASC data to inject the sentiment knowledge. The sentiment-oriented dependency tree is then induced from the fine-tuned PLMs, which further improves the performance of several state-of-the-art dependency-based models. 

So far, the common viewpoint in the NLP community is that PLMs are capable of accurately reflecting the semantic meanings of input words \citep{plm-survey-qxp}. 
However, the contextualized embeddings obtained via the self-attention mechanism that captures full word dependencies within the sentence are presumably redundant for the ABSA tasks. In fact, the superiority of the works~\citep{coling20-ate-constituency,acl20-relational,naacl21-asc-graphmerge}, which explicitly guide the further transformation of PLM representations with meaningful structure, to those using ``[CLS]'' representation for predictions indirectly suggests the existence of such redundancy. How to consolidate meaningful and sparse structure with PLMs or refine the intrinsic fully-connected self-attention for obtaining ABSA-related representations in a more efficient way deserves more attention and research efforts.
On the other hand, there is still room for improving the robustness of PLM-based ABSA models. Particularly, as observed in~\citet{emnlp20-asc-tasty}, even though the PLM-based models significantly outperform the previous neural models on the adversarial examples, it still suffers from more than 25\% performance drop on the simplest ASC task. 
We believe that exploiting PLMs for truly understanding the aspect-level sentiment, e.g., being robust to the reversed opinion and sentiment negation, instead of learning the spurious correlations between the aspect and sentiment labels is the future challenge for building PLM-based ABSA models. But there is still a long way to realize such kind of intelligence.

\section{Transferable ABSA}
\subsection{Cross-domain ABSA} \label{sec:cross-domain}
The supervised ABSA models within a single domain have been well developed. However, in real-world scenarios which involve texts from multiple or even unseen domains, it is likely that these models fail to obtain satisfactory predictions. The major reason is that the aspects referring to the opinion target from different domains are usually of great difference, and the models may not have prior knowledge about the frequently-used terms in the unseen domains. A straightforward solution is to create labeled data for these domains and re-train additional in-domain models. Considering that the ABSA tasks require fine-grained annotations, it is often expensive or even impossible to collect sufficient amount of labeled data. To enable the cross-domain ABSA predictions at lower cost, domain adaptation techniques~\citep{emnlp06-domain,tkde10-cross-domain-survey} are employed to provide alternative solutions for well generalizing the ABSA systems to other domains. Roughly speaking, the majority of these works can be separated into two groups: \textbf{feature-based transfer} and \textbf{data-based transfer}. 

The core idea of \textbf{feature-based transfer} is to learn domain-independent representations for the ABSA tasks. \citet{emnlp10-ate-extracting} and \citet{semeval14-ate-ihs} instantiate this idea by introducing rich syntactic features, which are invariant across domains, into a CRF tagger for the cross-domain ATE task. \citet{acl18-ate-recursive,cl19-ate-syntactically} design a dependency edge prediction task to enforce the learning of syntactic-aware representations, with the aim of reducing the domain shift at the word level. Other auxiliary tasks, such as domain classification~\citep{tac19-neural,tkde21-asc-eatn}, aspect-opinion interaction prediction~\citep{aaai19-ate-transferable} and opinion term detection~\citep{emnlp19-uabsa-transferable}, are also integrated to better align the representations in different domains. Different from the above studies, \citet{acl21-ate-bridge} simply aggregate the syntactic roles for each word and regard the syntactic embedding as the bridge between the source domain and target domain, which considerably improves the efficacy of domain adaptation. \citet{tnnls21-ate-weakly} assume the availability of sentence-level aspect category annotations in the target domain and propose an interaction transfer network to capture the domain-invariant category-term correlations. 

Compared to feature-based transfer, \textbf{data-based transfer} aims to adjust the distribution of the training data to better generalize the ABSA model to the target domain \citep{acl21-uabsa-cross, naacl22-xdomain}. \citet{aaai17-ate-recurrent} employ high-precision syntactic patterns together with some domain-independent opinion terms to create pseudo-labeled data in the target domain. The pseudo-labeled target data is then augmented to the source domain training set for building cross-domain ABSA models. \citet{acl12-ate-cross} build target-domain pseudo-labeled data in a similar way and perform re-weighting on the source domain training instances based on the pseudo-labeled data. Instead of producing supervision signals on the unlabeled target-domain data, \citet{acl21-uabsa-cross} develop an aspect-constrained (opinion-constrained) masked language model, which takes the source domain labeled reviews as input and perform in-place aspect term (opinion term) conversion from the source domain to the target domain as the silver training data.

In addition, \citet{emnlp20-uabsa-uf} propose to couple a token-level instance re-weighting strategy with domain-invariant representation learning from auxiliary tasks to consolidate feature-based transfer and data-based transfer for better domain adaptation for cross-domain E2E-ABSA.
\citet{coling20-ate-syntactically} and \citet{lrec20-adapt} regard the embeddings from PLMs as the features for the ABSA predictions across different domains and obtain reasonable results, showing that the PLMs pre-trained on large-scale corpus are already able to provide good domain-independent representations. \citet{emnlp20-uabsa-dombert} further strengthen the domain specificity of PLMs by continually pre-training BERT with the unlabeled texts from multiple relevant domains, drastically improving the domain generalization capability of BERT on the E2E-ABSA task. Such advances suggest that the consolidation of feature-based transfer and data-based transfer is a better way for cross-domain ABSA and language model pre-training can be introduced as a plug-and-play component to further enhance the domain adaptation performance.

\subsection{Cross-lingual ABSA} \label{sec:xabsa}
The majority of existing ABSA works are conducted on the resource-rich language (mostly in English), while opinions are often expressed in different languages in practice. However, annotating labeled data for each language can be time-consuming, which motivates the task of cross-lingual ABSA (XABSA). Due to the difficulty of cross-lingual transfer, most XABSA studies are conducted on simple ABSA tasks such as cross-lingual aspect term extraction (XATE) \citep{taslp15-xabsa-wxj, conll15-xabsa-instance, ijcai18-xabsa-transition, naacl19-xabsa-zs-ate}, cross-lingual aspect sentiment classification (XASC) \citep{acl15-xabsa-smt, coling16-xabsa-bilingual-embed, naacl18-xabsa-sparse}, and cross-lingual End-to-End ABSA \citep{arxiv20-xabsa-lx, emnlp21-xabsa-wxzhang}.

To realize the cross-lingual transfer, the key problem is to obtain the language-specific knowledge in the target language. 
Early methods typically rely on translation systems to obtain such knowledge. The sentence is first translated from the source to the target language with an off-the-shelf translation system. The label is then similarly projected from the source to target, either directly or with word alignment tools such as FastAlign \citep{fastalign} since some ABSA tasks (e.g., XATE) require token-level annotations. Therefore, an ABSA model can be trained with the obtained (pseudo-)labeled target language data. Because the performance of this type of method heavily relies on the quality of the translation and label projection, many techniques have been proposed to improve the data quality, including the co-training strategy \citep{taslp15-xabsa-wxj}, instance selection \citep{conll15-xabsa-instance}, or constrained SMT \citep{acl15-xabsa-smt}.

Cross-lingual word embeddings pre-trained on large parallel bilingual corpus have also been used for XABSA. By sharing a common vector space, the model can be used in a language-agnostic manner \citep{coling16-xabsa-bilingual-embed, naacl18-xabsa-sparse}. For example, \citet{ijcai18-xabsa-transition} utilize a transition-based mechanism to tackle the XATE task by aligning representations in different languages into a shared space through an adversarial network. \citet{naacl19-xabsa-zs-ate} consider the zero-shot ATE task with two types of cross-lingual word embeddings. Especially, they find that transferring from multiple source languages can largely improve the performance.

Recently, inspired by the success of exploiting monolingual PLMs, utilizing multilingual PLMs (mPLMs) such as multilingual BERT \citep{naacl19-bert} and XLM-RoBERTa \citep{acl20-xlmr} to tackle cross-lingual NLP tasks has become a common practice. Typically, a PLM is first pre-trained on a large volume of multilingual corpus, then fine-tuned on the source language data for learning task-specific knowledge. Finally, it can be directly used to conduct inference on the target language testing data (called \textit{zero-shot transfer}). Thanks to the language knowledge obtained in the pre-training stage, zero-shot transfer has shown to be an effective method for many cross-lingual NLP tasks \citep{acl19-mbert-study, iclr20-mbert-study}. However, the language knowledge learned in the pre-training step may be insufficient for the XABSA problem. As compensation, utilizing the translated (pseudo-)labeled target language data can equip the model with richer target language knowledge. For example, \citet{arxiv20-xabsa-lx} propose a warm-up mechanism to distill the knowledge from the translated data in each language to enhance the performance. \citet{emnlp21-xabsa-wxzhang} point out the importance of the translated target language data and propose an alignment-free label projection method to obtain high-quality pseudo-labeled target data. They show that even fine-tuning the mPLMs on such data can establish a strong baseline for the XABSA task.

Compared with the monolingual ABSA problem, the XABSA problem is relatively underexplored. Although mPLMs are widely used for various cross-lingual NLP tasks nowadays, exploring their usage in the XABSA can be tricky since language-specific knowledge plays an essential role in any ABSA task. Therefore, it calls for better adaption strategies of the mPLMs to inject the model with richer target language knowledge.
On the other hand, existing studies mainly focus on relatively easier ABSA tasks, exploring the cross-lingual transfer for more difficult compound ABSA tasks can be challenging while useful in practice.

\section{Challenges and Future Directions}
Over the last decade, we have seen great progress on the ABSA problem, either new tasks or novel methods. Despite the progress, there remains challenges for building more intelligent and powerful ABSA systems. In this section, we discuss some challenges, as well as potential directions which we hope can help advance the ABSA study.

\subsection{Quest for Larger and More Challenging Datasets}
As discussed in Sec \ref{sec:dataset}, most existing ABSA datasets are derived from the SemEval shared challenges~\citep{semeval14-absa, semeval15-absa, semeval16-absa} with additional data processing and annotations for specific tasks. However, the relatively small size of data (e.g., hundreds of sentences) makes it difficult to clearly compare different models, especially for PLM-based models having millions of parameters. Currently, it is a common practice to train a model with different random seeds (often five or ten) and report the model performance with averaged scores across different runs, but it would be better to introduce larger datasets for more fair and reliable comparisons.
Besides, although existing datasets provide valuable test beds for comparing different methods, there remains a great need for proposing more challenging datasets to satisfy the real-world scenarios. For example, datasets containing reviews from multiples domains or multiple languages can help evaluate multi-domain and multi-lingual ABSA systems. Moreover, since user opinions can be expressed in any kind of format, we also expect datasets collected from different opinion-sharing platforms such as question-answering platforms~\citep{emnlp21-findings-uabsa-qa} or customer service dialogs~\citep{aaai20-sentiment-dialog}.

\subsection{Multimodal ABSA}
Most existing ABSA works focus on analyzing opinionated texts such as customer reviews or tweets. However, users often share their opinions with other modalities such as images. Since contents in different modalities are often closely related, exploiting such multimodal information can help better analyze users' sentiments towards different aspects \citep{ijcai22-mmabsa, acl22-mmabsa}. Recent studies on multimodal ABSA mainly concentrate on simple ABSA tasks such as multimodal ATE~\citep{nlpcc20-mm-ate, neuralcomp21-mm-acsc} and multimodal ASC~\citep{aaai19-mm-acsc, ijcai19-mm-atsc, taslp20-mm-atsc, mm21-mm-atsc}. To align the information from different modalities, the text and image are often first encoded to feature representations, then some interaction networks are designed to fuse the information for making the final prediction.
More recently, inspired by the success of the E2E-ABSA task in a single modal (i.e., based on texts only), \citet{emnlp21-mm-e2e-absa} study the task of multimodal E2E-ABSA, aiming to capture the connection between its two subtasks in the multimodal scenario. They present a multimodal joint learning method with auxiliary cross-modal relation detection to obtain all aspect term and sentiment polarity pairs.
Despite these initial attempts, there remain some promising directions: from the perspective of the task, handling more complicated multimodal ABSA tasks should be considered; from the perspective of the method, more advanced multimodal techniques should be proposed for fusing the multimodal opinion information, e.g., constructing models based on the multimodal PLMs.
We believe multimodal ABSA would receive more attention given its increasing popularity in real-world applications.

\subsection{Unified Model for Multiple Tasks}
During the introduction of various ABSA tasks, we can notice that some ideas and model designs appear from time to time. Indeed, solutions to one ABSA task can be easily borrowed to tackle another similar task since these tasks are often closely related. This naturally posts a question: can we build a unified model that can tackle multiple (if not all) ABSA tasks at the same time? If so, there is no need to design specific models for every task. It can be also useful in practice because we may not want to change the model architecture and re-train it every time we have some new data with different types of opinion annotations. 
In Sec~\ref{sec:modeling-paradigm}, we show that different tasks can be tackled via the same model if they can be formulated as the same modeling paradigm. Several recent studies demonstrate some initial attempts along this direction. They either transform the task into the \texttt{MRC} paradigm by designing task-specific queries for the MRC model \citep{aaai21-aope-mrc, aaai21-aste-bimrc, aaai21-aste-dualmrc}, or the \texttt{Seq2Seq} paradigm by directly generating the target sentiment elements in the natural language form \citep{acl21-gabsa, acl21-aste-generative-qxp, emnlp21-quad}. 
Beyond solving multiple tasks with the same architecture, \citet{emnlp21-quad} further found that the task-specific knowledge could be easily transferred across different ABSA tasks (called \textit{cross-task transfer}) if they are under the same modeling paradigm.
We expect more research efforts would appear to enable more practically useful ABSA systems.

\subsection{Lifelong ABSA}\label{sec:lifelong}
Lifelong learning, also referred to as continual learning, aims at accumulating knowledge learned from previous tasks and adapting it for helping future learning during a sequence of tasks~\citep{delange2021continual}. 
\citet{acl15-lifelongsa} first study the sentiment analysis from the perspective of lifelong learning and propose the lifelong sentiment classification problem which requires a model to tackle a series of sentiment classification tasks.
\citet{bigdata18-lifelongabsa} impose the idea of lifelong learning into the ASC task with memory networks. 
Recent studies begin to investigate the catastrophic forgetting issue during the sequential learning~\citep{sigir21-lifelongsa,coling20-lifelongsa,emnlp21-continualabsa,naacl21-continualabsa}, instead of simply studying it as an extension of cross-domain sentiment analysis for knowledge accumulation. However, existing studies mainly focus on domain incremental learning for the ASC task~\citep{emnlp21-continualabsa,naacl21-continualabsa}, where all tasks sharing the same fixed label classes (e.g., positive, negative, and neutral) and no task information is required. 
To develop more advanced lifelong ABSA systems, it inevitably requires studying the incremental learning of the class and task. For instance, the classes of aspect categories vary in different applications, which calls for methods that can adapt to the changing categories. 
Besides, cross-task transfer~\citep{emnlp21-quad} has been shown to be effective in transferring knowledge learned from low-level ABSA tasks to high-level ABSA tasks. Therefore, it is also worth exploring lifelong learning across different types of ABSA tasks.

\section{Conclusions}
This survey aims to provide a comprehensive review of the aspect-based sentiment analysis problem, including its various tasks, methods, current challenges, and potential directions. We first set up the background of ABSA research with the four sentiment elements of ABSA, the definition, common modeling paradigms, and existing resources. Then we describe each ABSA task with their corresponding solutions in detail, with an emphasis on the recent advances of the compound ABSA tasks. Meanwhile, we categorize existing studies from the sentiment elements involved and summarize representative methods of different modeling paradigms for each task, which provides a clear picture of current progress. We further discuss the utilization of pre-trained language models for the ABSA problem, which has brought large improvements to a wide variety of ABSA tasks. We investigate the advantages they have, as well as their limitations. Besides, we review advances of cross-domain and cross-lingual ABSA, which can lead to more practical ABSA systems. Finally, we discuss some current challenges and promising future directions for this field.

\section*{Acknowledgments}
The work described in this paper is substantially supported by a grant from the Research Grant Council of the Hong Kong Special Administrative Region, China (Project Code: 14200719).

\ifCLASSOPTIONcaptionsoff
  \newpage
\fi

\bibliographystyle{IEEEtranN}
\bibliography{IEEEabrv}

\begin{thebibliography}{202}
\providecommand{\natexlab}[1]{#1}
\providecommand{\url}[1]{#1}
\csname url@samestyle\endcsname
\providecommand{\newblock}{\relax}
\providecommand{\bibinfo}[2]{#2}
\providecommand{\BIBentrySTDinterwordspacing}{\spaceskip=0pt\relax}
\providecommand{\BIBentryALTinterwordstretchfactor}{4}
\providecommand{\BIBentryALTinterwordspacing}{\spaceskip=\fontdimen2\font plus
\BIBentryALTinterwordstretchfactor\fontdimen3\font minus
  \fontdimen4\font\relax}
\providecommand{\BIBforeignlanguage}[2]{{%
\expandafter\ifx\csname l@#1\endcsname\relax
\typeout{** WARNING: IEEEtranN.bst: No hyphenation pattern has been}%
\typeout{** loaded for the language `#1'. Using the pattern for}%
\typeout{** the default language instead.}%
\else
\language=\csname l@#1\endcsname
\fi
#2}}
\providecommand{\BIBdecl}{\relax}
\BIBdecl

\bibitem[Liu(2012)]{hlt12-liu-bing-sa}
B.~Liu, ``Sentiment analysis and opinion mining,'' \emph{Synthesis lectures on
  human language technologies}, vol.~5, no.~1, pp. 1--167, 2012.

\bibitem[Turney(2002)]{acl02-thumbs}
P.~D. Turney, ``Thumbs up or thumbs down? semantic orientation applied to
  unsupervised classification of reviews,'' in \emph{ACL}, 2002, pp. 417--424.

\bibitem[Pang et~al.(2002)Pang, Lee, and Vaithyanathan]{emnlp02-pang-thumbs}
B.~Pang, L.~Lee, and S.~Vaithyanathan, ``Thumbs up? sentiment classification
  using machine learning techniques,'' in \emph{EMNLP}, 2002, pp. 79--86.

\bibitem[Yu and Hatzivassiloglou(2003)]{emnlp02-yu-sa}
H.~Yu and V.~Hatzivassiloglou, ``Towards answering opinion questions:
  Separating facts from opinions and identifying the polarity of opinion
  sentences,'' in \emph{EMNLP}, 2003, pp. 129--136.

\bibitem[Schouten and Frasincar(2016)]{tkde16-absa-survey}
K.~Schouten and F.~Frasincar, ``Survey on aspect-level sentiment analysis,''
  \emph{{IEEE} Trans. Knowl. Data Eng.}, vol.~28, no.~3, pp. 813--830, 2016.

\bibitem[Nazir et~al.(2020)Nazir, Rao, Wu, and Sun]{tac20-absa-survey}
A.~Nazir, Y.~Rao, L.~Wu, and L.~Sun, ``Issues and challenges of aspect-based
  sentiment analysis: A comprehensive survey,'' \emph{{IEEE} Trans. Affect.
  Comput.}, 2020.

\bibitem[Zhang et~al.(2021{\natexlab{a}})Zhang, Deng, Li, Yuan, Bing, and
  Lam]{emnlp21-quad}
W.~Zhang, Y.~Deng, X.~Li, Y.~Yuan, L.~Bing, and W.~Lam, ``Aspect sentiment quad
  prediction as paraphrase generation,'' in \emph{EMNLP}, 2021, pp. 9209--9219.

\bibitem[Liu et~al.(2015)Liu, Joty, and Meng]{emnlp15-ate-rnn}
P.~Liu, S.~R. Joty, and H.~M. Meng, ``Fine-grained opinion mining with
  recurrent neural networks and word embeddings,'' in \emph{EMNLP}, 2015, pp.
  1433--1443.

\bibitem[Jiang et~al.(2011)Jiang, Yu, Zhou, Liu, and Zhao]{acl11-asc-jiang}
L.~Jiang, M.~Yu, M.~Zhou, X.~Liu, and T.~Zhao, ``Target-dependent twitter
  sentiment classification,'' in \emph{ACL}, 2011, pp. 151--160.

\bibitem[Mitchell et~al.(2013)Mitchell, Aguilar, Wilson, and
  Durme]{emnlp13-uabsa}
M.~Mitchell, J.~Aguilar, T.~Wilson, and B.~V. Durme, ``Open domain targeted
  sentiment,'' in \emph{EMNLP}, 2013, pp. 1643--1654.

\bibitem[Chen et~al.(2020{\natexlab{a}})Chen, Liu, Wang, Zhang, and
  Chi]{acl20-aope-sync}
S.~Chen, J.~Liu, Y.~Wang, W.~Zhang, and Z.~Chi, ``Synchronous double-channel
  recurrent network for aspect-opinion pair extraction,'' in \emph{ACL}, 2020,
  pp. 6515--6524.

\bibitem[Peng et~al.(2020)Peng, Xu, Bing, Huang, Lu, and Si]{aaai20-aste}
H.~Peng, L.~Xu, L.~Bing, F.~Huang, W.~Lu, and L.~Si, ``Knowing what, how and
  why: {A} near complete solution for aspect-based sentiment analysis,'' in
  \emph{AAAI}, 2020, pp. 8600--8607.

\bibitem[Zhao et~al.(2020)Zhao, Huang, Zhang, Lu, and Xue]{acl20-aope-spanmlt}
H.~Zhao, L.~Huang, R.~Zhang, Q.~Lu, and H.~Xue, ``Spanmlt: {A} span-based
  multi-task learning framework for pair-wise aspect and opinion terms
  extraction,'' in \emph{ACL}, 2020, pp. 3239--3248.

\bibitem[Zhou et~al.(2019)Zhou, Huang, Chen, Hu, Wang, and
  He]{access19-absa-survey}
J.~Zhou, J.~X. Huang, Q.~Chen, Q.~V. Hu, T.~Wang, and L.~He, ``Deep learning
  for aspect-level sentiment classification: Survey, vision, and challenges,''
  \emph{{IEEE} Access}, vol.~7, pp. 78\,454--78\,483, 2019.

\bibitem[Liu et~al.(2020)Liu, Chatterjee, Zhou, Lu, and
  Abusorrah]{tcss20-absa-survey}
H.~Liu, I.~Chatterjee, M.~Zhou, X.~S. Lu, and A.~Abusorrah, ``Aspect-based
  sentiment analysis: {A} survey of deep learning methods,'' \emph{{IEEE}
  Trans. Comput. Soc. Syst.}, vol.~7, no.~6, pp. 1358--1375, 2020.

\bibitem[Poria et~al.(2020)Poria, Hazarika, Majumder, and
  Mihalcea]{tac20-sa-survey}
S.~Poria, D.~Hazarika, N.~Majumder, and R.~Mihalcea, ``Beneath the tip of the
  iceberg: Current challenges and new directions in sentiment analysis
  research,'' \emph{{IEEE} Trans. Affect. Comput.}, 2020.

\bibitem[Devlin et~al.(2019)Devlin, Chang, Lee, and Toutanova]{naacl19-bert}
J.~Devlin, M.~Chang, K.~Lee, and K.~Toutanova, ``{BERT:} pre-training of deep
  bidirectional transformers for language understanding,'' in \emph{NAACL-HLT},
  2019, pp. 4171--4186.

\bibitem[Liu et~al.(2019)Liu, Ott, Goyal, Du, Joshi, Chen, Levy, Lewis,
  Zettlemoyer, and Stoyanov]{arxiv19-roberta}
Y.~Liu, M.~Ott, N.~Goyal, J.~Du, M.~Joshi, D.~Chen, O.~Levy, M.~Lewis,
  L.~Zettlemoyer, and V.~Stoyanov, ``Roberta: {A} robustly optimized {BERT}
  pretraining approach,'' \emph{CoRR}, vol. abs/1907.11692, 2019.

\bibitem[Li et~al.(2019{\natexlab{a}})Li, Bing, Zhang, and
  Lam]{wnut19-exploiting}
X.~Li, L.~Bing, W.~Zhang, and W.~Lam, ``Exploiting {BERT} for end-to-end
  aspect-based sentiment analysis,'' in \emph{W-NUT}, 2019, pp. 34--41.

\bibitem[Pan and Yang(2010)]{tkde10-cross-domain-survey}
S.~J. Pan and Q.~Yang, ``A survey on transfer learning,'' \emph{{IEEE} Trans.
  Knowl. Data Eng.}, vol.~22, no.~10, pp. 1345--1359, 2010.

\bibitem[Ruder et~al.(2019)Ruder, Vulic, and
  S{\o}gaard]{jair19-cross-lingual-survey}
S.~Ruder, I.~Vulic, and A.~S{\o}gaard, ``A survey of cross-lingual word
  embedding models,'' \emph{J. Artif. Intell. Res.}, vol.~65, pp. 569--631,
  2019.

\bibitem[Yin et~al.(2016)Yin, Wei, Dong, Xu, Zhang, and
  Zhou]{ijcai16-ate-wordembed}
Y.~Yin, F.~Wei, L.~Dong, K.~Xu, M.~Zhang, and M.~Zhou, ``Unsupervised word and
  dependency path embeddings for aspect term extraction,'' in \emph{IJCAI},
  2016, pp. 2979--2985.

\bibitem[Wang et~al.(2016{\natexlab{a}})Wang, Pan, Dahlmeier, and
  Xiao]{emnlp16-ate-rncrf}
W.~Wang, S.~J. Pan, D.~Dahlmeier, and X.~Xiao, ``Recursive neural conditional
  random fields for aspect-based sentiment analysis,'' in \emph{EMNLP}, 2016,
  pp. 616--626.

\bibitem[Li and Lam(2017)]{emnlp17-ate-min}
X.~Li and W.~Lam, ``Deep multi-task learning for aspect term extraction with
  memory interaction,'' in \emph{EMNLP}, 2017, pp. 2886--2892.

\bibitem[Wang et~al.(2017)Wang, Pan, Dahlmeier, and Xiao]{aaai17-ate-cmla}
W.~Wang, S.~J. Pan, D.~Dahlmeier, and X.~Xiao, ``Coupled multi-layer attentions
  for co-extraction of aspect and opinion terms,'' in \emph{AAAI}, 2017, pp.
  3316--3322.

\bibitem[Li et~al.(2018{\natexlab{a}})Li, Bing, Li, Lam, and
  Yang]{ijcai18-ate-hast}
X.~Li, L.~Bing, P.~Li, W.~Lam, and Z.~Yang, ``Aspect term extraction with
  history attention and selective transformation,'' in \emph{IJCAI}, 2018, pp.
  4194--4200.

\bibitem[Xu et~al.(2018)Xu, Liu, Shu, and Yu]{acl18-ate-decnn}
H.~Xu, B.~Liu, L.~Shu, and P.~S. Yu, ``Double embeddings and cnn-based sequence
  labeling for aspect extraction,'' in \emph{ACL}, 2018, pp. 592--598.

\bibitem[Ma et~al.(2019)Ma, Li, Wu, Xie, and Wang]{acl19-ate-seq2label}
D.~Ma, S.~Li, F.~Wu, X.~Xie, and H.~Wang, ``Exploring sequence-to-sequence
  learning in aspect term extraction,'' in \emph{ACL}, 2019, pp. 3538--3547.

\bibitem[Yang et~al.(2020)Yang, Li, Quan, Shen, and
  Su]{coling20-ate-constituency}
Y.~Yang, K.~Li, X.~Quan, W.~Shen, and Q.~Su, ``Constituency lattice encoding
  for aspect term extraction,'' in \emph{COLING}, 2020, pp. 844--855.

\bibitem[Yin et~al.(2020)Yin, Wang, and Zhang]{coling20-ate-pod}
Y.~Yin, C.~Wang, and M.~Zhang, ``{P}o{D}: Positional dependency-based word
  embedding for aspect term extraction,'' in \emph{COLING}, 2020, pp.
  1714--1719.

\bibitem[Li et~al.(2020{\natexlab{a}})Li, Chen, Quan, Ling, and
  Song]{acl20-ate-maskedseq2seq}
K.~Li, C.~Chen, X.~Quan, Q.~Ling, and Y.~Song, ``Conditional augmentation for
  aspect term extraction via masked sequence-to-sequence generation,'' in
  \emph{ACL}, 2020, pp. 7056--7066.

\bibitem[Chen and Qian(2020{\natexlab{a}})]{emnlp20-ate-softproto}
Z.~Chen and T.~Qian, ``Enhancing aspect term extraction with soft prototypes,''
  in \emph{EMNLP}, 2020, pp. 2107--2117.

\bibitem[Wang et~al.(2021{\natexlab{a}})Wang, Wen, Zhao, Yang, and
  Xu]{emnlp21-ate-selftrain}
Q.~Wang, Z.~Wen, Q.~Zhao, M.~Yang, and R.~Xu, ``Progressive self-training with
  discriminator for aspect term extraction,'' in \emph{EMNLP}, 2021, pp.
  257--268.

\bibitem[He et~al.(2017)He, Lee, Ng, and Dahlmeier]{acl17-ruidan-ate}
R.~He, W.~S. Lee, H.~T. Ng, and D.~Dahlmeier, ``An unsupervised neural
  attention model for aspect extraction,'' in \emph{ACL}, 2017, pp. 388--397.

\bibitem[Luo et~al.(2019{\natexlab{a}})Luo, Ao, Song, Li, Yang, He, and
  Yu]{ijcai19-unsupervised-ate}
L.~Luo, X.~Ao, Y.~Song, J.~Li, X.~Yang, Q.~He, and D.~Yu, ``Unsupervised neural
  aspect extraction with sememes,'' in \emph{IJCAI}, 2019, pp. 5123--5129.

\bibitem[Liao et~al.(2019)Liao, Li, Zhang, Wang, Wu, and
  Wong]{emnlp19-ate-coupling}
M.~Liao, J.~Li, H.~Zhang, L.~Wang, X.~Wu, and K.-F. Wong, ``Coupling global and
  local context for unsupervised aspect extraction,'' in \emph{EMNLP-IJCNLP},
  2019, pp. 4579--4589.

\bibitem[Zhou et~al.(2015{\natexlab{a}})Zhou, Wan, and Xiao]{aaai15-acd-zhou}
X.~Zhou, X.~Wan, and J.~Xiao, ``Representation learning for aspect category
  detection in online reviews,'' in \emph{AAAI}, 2015, pp. 417--424.

\bibitem[Movahedi et~al.(2019)Movahedi, Ghadery, Faili, and
  Shakery]{arxiv19-acd-topic}
S.~Movahedi, E.~Ghadery, H.~Faili, and A.~Shakery, ``Aspect category detection
  via topic-attention network,'' \emph{CoRR}, vol. abs/1901.01183, 2019.

\bibitem[Ghadery et~al.(2019)Ghadery, Movahedi, Sabet, Faili, and
  Shakery]{ecir19-acd-matching}
E.~Ghadery, S.~Movahedi, M.~J. Sabet, H.~Faili, and A.~Shakery, ``{LICD:} {A}
  language-independent approach for aspect category detection,'' in
  \emph{ECIR}, 2019, pp. 575--589.

\bibitem[Hu et~al.(2021)Hu, Zhao, Guo, Xue, Gao, Gao, Cheng, and
  Su]{acl21-acd-multi-label}
M.~Hu, S.~Zhao, H.~Guo, C.~Xue, H.~Gao, T.~Gao, R.~Cheng, and Z.~Su,
  ``Multi-label few-shot learning for aspect category detection,'' in
  \emph{ACL-IJCNLP}, 2021, pp. 6330--6340.

\bibitem[Tulkens and van Cranenburgh(2020)]{acl20-acd-cat}
S.~Tulkens and A.~van Cranenburgh, ``Embarrassingly simple unsupervised aspect
  extraction,'' in \emph{ACL}, 2020, pp. 3182--3187.

\bibitem[Shi et~al.(2021)Shi, Li, Wang, and Reddy]{aaai21-acd}
T.~Shi, L.~Li, P.~Wang, and C.~K. Reddy, ``A simple and effective
  self-supervised contrastive learning framework for aspect detection,'' in
  \emph{{AAAI}}, 2021, pp. 13\,815--13\,824.

\bibitem[Yu et~al.(2019)Yu, Jiang, and Xia]{taslp19-aoce-yu}
J.~Yu, J.~Jiang, and R.~Xia, ``Global inference for aspect and opinion terms
  co-extraction based on multi-task neural networks,'' \emph{{IEEE} {ACM}
  Trans. Audio Speech Lang. Process.}, vol.~27, no.~1, pp. 168--177, 2019.

\bibitem[Wu et~al.(2020{\natexlab{a}})Wu, Wang, and Pan]{emnlp20-aoce-wwy}
M.~Wu, W.~Wang, and S.~J. Pan, ``Deep weighted maxsat for aspect-based opinion
  extraction,'' in \emph{EMNLP}, 2020, pp. 5618--5628.

\bibitem[Fan et~al.(2019)Fan, Wu, Dai, Huang, and Chen]{naacl19-towe}
Z.~Fan, Z.~Wu, X.~Dai, S.~Huang, and J.~Chen, ``Target-oriented opinion words
  extraction with target-fused neural sequence labeling,'' in \emph{NAACL-HLT},
  2019, pp. 2509--2518.

\bibitem[Wan et~al.(2020)Wan, Yang, Du, Liu, Qi, and Pan]{aaai20-tasd}
H.~Wan, Y.~Yang, J.~Du, Y.~Liu, K.~Qi, and J.~Z. Pan, ``Target-aspect-sentiment
  joint detection for aspect-based sentiment analysis,'' in \emph{AAAI}, 2020,
  pp. 9122--9129.

\bibitem[Veyseh et~al.(2020)Veyseh, Nouri, Dernoncourt, Dou, and
  Nguyen]{emnlp20-towe-syntax}
A.~P.~B. Veyseh, N.~Nouri, F.~Dernoncourt, D.~Dou, and T.~H. Nguyen,
  ``Introducing syntactic structures into target opinion word extraction with
  deep learning,'' in \emph{EMNLP}, 2020, pp. 8947--8956.

\bibitem[Mensah et~al.(2021)Mensah, Sun, and Aletras]{emnlp21-towe-position}
S.~Mensah, K.~Sun, and N.~Aletras, ``An empirical study on leveraging position
  embeddings for target-oriented opinion words extraction,'' in \emph{EMNLP},
  2021, pp. 9174--9179.

\bibitem[Dong et~al.(2014)Dong, Wei, Tan, Tang, Zhou, and Xu]{acl14-asc-dong}
L.~Dong, F.~Wei, C.~Tan, D.~Tang, M.~Zhou, and K.~Xu, ``Adaptive recursive
  neural network for target-dependent twitter sentiment classification,'' in
  \emph{ACL}, 2014, pp. 49--54.

\bibitem[Vo and Zhang(2015)]{ijcai15-asc-nn}
D.~Vo and Y.~Zhang, ``Target-dependent twitter sentiment classification with
  rich automatic features,'' in \emph{IJCAI}, 2015, pp. 1347--1353.

\bibitem[Tang et~al.(2016{\natexlab{a}})Tang, Qin, Feng, and
  Liu]{coling16-asc-effective-lstm}
D.~Tang, B.~Qin, X.~Feng, and T.~Liu, ``Effective lstms for target-dependent
  sentiment classification,'' in \emph{COLING}, 2016, pp. 3298--3307.

\bibitem[Wang et~al.(2016{\natexlab{b}})Wang, Huang, Zhu, and
  Zhao]{emnlp16-asc-atae}
Y.~Wang, M.~Huang, X.~Zhu, and L.~Zhao, ``Attention-based {LSTM} for
  aspect-level sentiment classification,'' in \emph{EMNLP}, 2016, pp. 606--615.

\bibitem[Tang et~al.(2016{\natexlab{b}})Tang, Qin, and Liu]{emnlp16-asc-memory}
D.~Tang, B.~Qin, and T.~Liu, ``Aspect level sentiment classification with deep
  memory network,'' in \emph{EMNLP}, 2016, pp. 214--224.

\bibitem[Ma et~al.(2017)Ma, Li, Zhang, and Wang]{ijcai17-asc-ian}
D.~Ma, S.~Li, X.~Zhang, and H.~Wang, ``Interactive attention networks for
  aspect-level sentiment classification,'' in \emph{IJCAI}, 2017, pp.
  4068--4074.

\bibitem[Liu and Zhang(2017)]{eacl17-asc-attention}
J.~Liu and Y.~Zhang, ``Attention modeling for targeted sentiment,'' in
  \emph{EACL}, 2017, pp. 572--577.

\bibitem[Chen et~al.(2017)Chen, Sun, Bing, and Yang]{emnlp17-asc-ram}
P.~Chen, Z.~Sun, L.~Bing, and W.~Yang, ``Recurrent attention network on memory
  for aspect sentiment analysis,'' in \emph{EMNLP}, 2017, pp. 452--461.

\bibitem[Cheng et~al.(2017)Cheng, Zhao, Zhang, King, Zhang, and
  Wang]{cikm17-asc-heat}
J.~Cheng, S.~Zhao, J.~Zhang, I.~King, X.~Zhang, and H.~Wang, ``Aspect-level
  sentiment classification with {HEAT} (hierarchical attention) network,'' in
  \emph{ACM CIKM}, 2017, pp. 97--106.

\bibitem[Xue and Li(2018)]{acl18-asc-cnn}
W.~Xue and T.~Li, ``Aspect based sentiment analysis with gated convolutional
  networks,'' in \emph{ACL}, 2018, pp. 2514--2523.

\bibitem[Li et~al.(2018{\natexlab{b}})Li, Bing, Lam, and
  Shi]{acl18-asc-lx-tnet}
X.~Li, L.~Bing, W.~Lam, and B.~Shi, ``Transformation networks for
  target-oriented sentiment classification,'' in \emph{ACL}, 2018, pp.
  946--956.

\bibitem[Fan et~al.(2018)Fan, Feng, and Zhao]{emnlp18-asc-mgan}
F.~Fan, Y.~Feng, and D.~Zhao, ``Multi-grained attention network for
  aspect-level sentiment classification,'' in \emph{EMNLP}, 2018, pp.
  3433--3442.

\bibitem[Zhang et~al.(2019{\natexlab{a}})Zhang, Li, and Song]{sigir19-asc-pwcn}
C.~Zhang, Q.~Li, and D.~Song, ``Syntax-aware aspect-level sentiment
  classification with proximity-weighted convolution network,'' in
  \emph{SIGIR}, 2019, pp. 1145--1148.

\bibitem[Xu et~al.(2019{\natexlab{a}})Xu, Liu, Shu, and Yu]{naacl19-bertabsa}
H.~Xu, B.~Liu, L.~Shu, and P.~Yu, ``{BERT} post-training for review reading
  comprehension and aspect-based sentiment analysis,'' in \emph{NAACL-HLT},
  2019, pp. 2324--2335.

\bibitem[Chen and Qian(2019)]{acl19-asc-transcap}
Z.~Chen and T.~Qian, ``Transfer capsule network for aspect level sentiment
  classification,'' in \emph{ACL}, 2019, pp. 547--556.

\bibitem[Du et~al.(2019)Du, Sun, Wang, Qi, Liao, Xu, and
  Liu]{emnlp19-asc-capsule}
C.~Du, H.~Sun, J.~Wang, Q.~Qi, J.~Liao, T.~Xu, and M.~Liu, ``Capsule network
  with interactive attention for aspect-level sentiment classification,'' in
  \emph{EMNLP-IJCNLP}, 2019, pp. 5489--5498.

\bibitem[Liang et~al.(2019)Liang, Meng, Zhang, Xu, Chen, and
  Zhou]{emnlp19-asc-agdt}
Y.~Liang, F.~Meng, J.~Zhang, J.~Xu, Y.~Chen, and J.~Zhou, ``A novel
  aspect-guided deep transition model for aspect based sentiment analysis,'' in
  \emph{EMNLP-IJCNLP}, 2019, pp. 5569--5580.

\bibitem[Zhang et~al.(2019{\natexlab{b}})Zhang, Li, and Song]{emnlp19-asgcn}
C.~Zhang, Q.~Li, and D.~Song, ``Aspect-based sentiment classification with
  aspect-specific graph convolutional networks,'' in \emph{EMNLP-IJCNLP}, 2019,
  pp. 4568--4578.

\bibitem[Sun et~al.(2019{\natexlab{a}})Sun, Zhang, Mensah, Mao, and
  Liu]{emnlp19-asc-cdt}
K.~Sun, R.~Zhang, S.~Mensah, Y.~Mao, and X.~Liu, ``Aspect-level sentiment
  analysis via convolution over dependency tree,'' in \emph{EMNLP-IJCNLP},
  2019, pp. 5678--5687.

\bibitem[Tang et~al.(2020)Tang, Ji, Li, and Zhou]{acl20-asc-dgedt}
H.~Tang, D.~Ji, C.~Li, and Q.~Zhou, ``Dependency graph enhanced
  dual-transformer structure for aspect-based sentiment classification,'' in
  \emph{ACL}, 2020, pp. 6578--6588.

\bibitem[Chen et~al.(2020{\natexlab{b}})Chen, Teng, and
  Zhang]{emnlp20-latent-graph}
C.~Chen, Z.~Teng, and Y.~Zhang, ``Inducing target-specific latent structures
  for aspect sentiment classification,'' in \emph{EMNLP}, 2020, pp. 5596--5607.

\bibitem[Xu et~al.(2020{\natexlab{a}})Xu, Bing, Lu, and
  Huang]{emnlp20-asc-xulu}
L.~Xu, L.~Bing, W.~Lu, and F.~Huang, ``Aspect sentiment classification with
  aspect-specific opinion spans,'' in \emph{EMNLP}, 2020, pp. 3561--3567.

\bibitem[Hou et~al.(2021)Hou, Qi, Wang, Ying, Huang, He, and
  Zhou]{naacl21-asc-graphmerge}
X.~Hou, P.~Qi, G.~Wang, R.~Ying, J.~Huang, X.~He, and B.~Zhou, ``Graph ensemble
  learning over multiple dependency trees for aspect-level sentiment
  classification,'' in \emph{NAACL-HLT}, 2021, pp. 2884--2894.

\bibitem[Tian et~al.(2021)Tian, Chen, and Song]{naacl21-tgcn}
Y.~Tian, G.~Chen, and Y.~Song, ``Aspect-based sentiment analysis with
  type-aware graph convolutional networks and layer ensemble,'' in
  \emph{NAACL-HLT}, 2021, pp. 2910--2922.

\bibitem[Li et~al.(2021)Li, Chen, Feng, Ma, Wang, and Hovy]{acl21-asc-dualgcn}
R.~Li, H.~Chen, F.~Feng, Z.~Ma, X.~Wang, and E.~H. Hovy, ``Dual graph
  convolutional networks for aspect-based sentiment analysis,'' in
  \emph{ACL-IJCNLP}, 2021, pp. 6319--6329.

\bibitem[Wang et~al.(2021{\natexlab{b}})Wang, Shen, Long, Zhou, and
  Chang]{emnlp21-asc-eliminating}
B.~Wang, T.~Shen, G.~Long, T.~Zhou, and Y.~Chang, ``Eliminating sentiment bias
  for aspect-level sentiment classification with unsupervised opinion
  extraction,'' in \emph{Findings of EMNLP}, 2021, pp. 3002--3012.

\bibitem[Zhou et~al.(2021{\natexlab{a}})Zhou, Liao, Gao, Jie, and
  Lu]{emnlp21-asc-aclt}
Y.~Zhou, L.~Liao, Y.~Gao, Z.~Jie, and W.~Lu, ``To be closer: Learning to link
  up aspects with opinions,'' in \emph{EMNLP}, 2021, pp. 3899--3909.

\bibitem[Wu et~al.(2020{\natexlab{b}})Wu, Ying, Zhao, Fan, Dai, and
  Xia]{emnlp20-aste-grid}
Z.~Wu, C.~Ying, F.~Zhao, Z.~Fan, X.~Dai, and R.~Xia, ``Grid tagging scheme for
  aspect-oriented fine-grained opinion extraction,'' in \emph{Findings of
  EMNLP}, 2020, pp. 2576--2585.

\bibitem[Gao et~al.(2021)Gao, Wang, Liu, Wang, Zhang, and
  Liao]{aaai21-aope-mrc}
L.~Gao, Y.~Wang, T.~Liu, J.~Wang, L.~Zhang, and J.~Liao, ``Question-driven span
  labeling model for aspect-opinion pair extraction,'' in \emph{AAAI}, 2021,
  pp. 12\,875--12\,883.

\bibitem[Wu et~al.(2021{\natexlab{a}})Wu, Fei, Ren, Ji, and
  Li]{ijcai21-aope-syntax}
S.~Wu, H.~Fei, Y.~Ren, D.~Ji, and J.~Li, ``Learn from syntax: Improving
  pair-wise aspect and opinion terms extraction with rich syntactic
  knowledge,'' in \emph{IJCAI}, 2021, pp. 3957--3963.

\bibitem[Zhang et~al.(2015)Zhang, Zhang, and Vo]{emnlp15-uabsa}
M.~Zhang, Y.~Zhang, and D.~Vo, ``Neural networks for open domain targeted
  sentiment,'' in \emph{EMNLP}, 2015, pp. 612--621.

\bibitem[Wang et~al.(2018{\natexlab{a}})Wang, Lan, and Wang]{ijcnn18-uabsa}
F.~Wang, M.~Lan, and W.~Wang, ``Towards a one-stop solution to both aspect
  extraction and sentiment analysis tasks with neural multi-task learning,'' in
  \emph{IJCNN}, 2018, pp. 1--8.

\bibitem[Li et~al.(2019{\natexlab{b}})Li, Bing, Li, and Lam]{aaai19-uabsa-lx}
X.~Li, L.~Bing, P.~Li, and W.~Lam, ``A unified model for opinion target
  extraction and target sentiment prediction,'' in \emph{AAAI}, 2019, pp.
  6714--6721.

\bibitem[Luo et~al.(2019{\natexlab{b}})Luo, Li, Liu, and
  Zhang]{acl19-uabsa-doer}
H.~Luo, T.~Li, B.~Liu, and J.~Zhang, ``{DOER:} dual cross-shared {RNN} for
  aspect term-polarity co-extraction,'' in \emph{ACL}, 2019, pp. 591--601.

\bibitem[He et~al.(2019)He, Lee, Ng, and Dahlmeier]{acl19-uabsa-imn}
R.~He, W.~S. Lee, H.~T. Ng, and D.~Dahlmeier, ``An interactive multi-task
  learning network for end-to-end aspect-based sentiment analysis,'' in
  \emph{ACL}, 2019, pp. 504--515.

\bibitem[Hu et~al.(2019{\natexlab{a}})Hu, Peng, Huang, Li, and
  Lv]{acl19-openabsa}
M.~Hu, Y.~Peng, Z.~Huang, D.~Li, and Y.~Lv, ``Open-domain targeted sentiment
  analysis via span-based extraction and classification,'' in \emph{ACL}, 2019,
  pp. 537--546.

\bibitem[Chen and Qian(2020{\natexlab{b}})]{acl20-uabsa-racl}
Z.~Chen and T.~Qian, ``Relation-aware collaborative learning for unified
  aspect-based sentiment analysis,'' in \emph{ACL}, 2020, pp. 3685--3694.

\bibitem[Luo et~al.(2020)Luo, Ji, Li, Jiang, and Duan]{emnlp20-uabsa-grace}
H.~Luo, L.~Ji, T.~Li, D.~Jiang, and N.~Duan, ``{GRACE}: Gradient harmonized and
  cascaded labeling for aspect-based sentiment analysis,'' in \emph{Findings of
  EMNLP}, 2020, pp. 54--64.

\bibitem[Liang et~al.(2021{\natexlab{a}})Liang, Meng, Zhang, Chen, Xu, and
  Zhou]{emnlp21-uabsa-iterative}
Y.~Liang, F.~Meng, J.~Zhang, Y.~Chen, J.~Xu, and J.~Zhou, ``An iterative
  multi-knowledge transfer network for aspect-based sentiment analysis,'' in
  \emph{Findings of EMNLP}, 2021, pp. 1768--1780.

\bibitem[Yu et~al.(2021{\natexlab{a}})Yu, Li, Luo, Meng, Ao, and
  He]{emnlp21-uabsa-self}
G.~Yu, J.~Li, L.~Luo, Y.~Meng, X.~Ao, and Q.~He, ``Self question-answering:
  Aspect-based sentiment analysis by role flipped machine reading
  comprehension,'' in \emph{Findings of EMNLP}, 2021, pp. 1331--1342.

\bibitem[Schmitt et~al.(2018)Schmitt, Steinheber, Schreiber, and
  Roth]{emnlp18-acsa-joint}
M.~Schmitt, S.~Steinheber, K.~Schreiber, and B.~Roth, ``Joint aspect and
  polarity classification for aspect-based sentiment analysis with end-to-end
  neural networks,'' in \emph{EMNLP}, 2018, pp. 1109--1114.

\bibitem[Cai et~al.(2020)Cai, Tu, Zhou, Yu, and Xia]{coling20-acsa-hiergcn}
H.~Cai, Y.~Tu, X.~Zhou, J.~Yu, and R.~Xia, ``Aspect-category based sentiment
  analysis with hierarchical graph convolutional network,'' in \emph{COLING},
  2020, pp. 833--843.

\bibitem[Liu et~al.(2021)Liu, Teng, Cui, Liu, and Zhang]{emnlp21-acsa-seq2seq}
J.~Liu, Z.~Teng, L.~Cui, H.~Liu, and Y.~Zhang, ``Solving aspect category
  sentiment analysis as a text generation task,'' in \emph{EMNLP}, 2021, pp.
  4406--4416.

\bibitem[Xu et~al.(2020{\natexlab{b}})Xu, Li, Lu, and
  Bing]{emnlp20-aste-position}
L.~Xu, H.~Li, W.~Lu, and L.~Bing, ``Position-aware tagging for aspect sentiment
  triplet extraction,'' in \emph{EMNLP}, 2020, pp. 2339--2349.

\bibitem[Zhang et~al.(2020)Zhang, Li, Song, and Wang]{emnlp20-aste-mtl}
C.~Zhang, Q.~Li, D.~Song, and B.~Wang, ``A multi-task learning framework for
  opinion triplet extraction,'' in \emph{Findings of EMNLP}, 2020, pp.
  819--828.

\bibitem[Chen et~al.(2021)Chen, Wang, Liu, and Wang]{aaai21-aste-bimrc}
S.~Chen, Y.~Wang, J.~Liu, and Y.~Wang, ``Bidirectional machine reading
  comprehension for aspect sentiment triplet extraction,'' in \emph{AAAI},
  2021, pp. 12\,666--12\,674.

\bibitem[Mao et~al.(2021)Mao, Shen, Yu, and Cai]{aaai21-aste-dualmrc}
Y.~Mao, Y.~Shen, C.~Yu, and L.~Cai, ``A joint training dual-mrc framework for
  aspect based sentiment analysis,'' in \emph{AAAI}, 2021, pp.
  13\,543--13\,551.

\bibitem[Zhang et~al.(2021{\natexlab{b}})Zhang, Li, Deng, Bing, and
  Lam]{acl21-gabsa}
W.~Zhang, X.~Li, Y.~Deng, L.~Bing, and W.~Lam, ``Towards generative
  aspect-based sentiment analysis,'' in \emph{ACL-IJCNLP}, 2021, pp. 504--510.

\bibitem[Yan et~al.(2021)Yan, Dai, Ji, Qiu, and
  Zhang]{acl21-aste-generative-qxp}
H.~Yan, J.~Dai, T.~Ji, X.~Qiu, and Z.~Zhang, ``A unified generative framework
  for aspect-based sentiment analysis,'' in \emph{ACL-IJCNLP}, 2021, pp.
  2416--2429.

\bibitem[Xu et~al.(2021)Xu, Chia, and Bing]{acl21-aste-span}
L.~Xu, Y.~K. Chia, and L.~Bing, ``Learning span-level interactions for aspect
  sentiment triplet extraction,'' in \emph{ACL-IJCNLP}, 2021, pp. 4755--4766.

\bibitem[Fei et~al.(2021)Fei, Ren, Zhang, and
  Ji]{tnnls21-aste-nonautoregressive}
H.~Fei, Y.~Ren, Y.~Zhang, and D.~Ji, ``Nonautoregressive encoder-decoder neural
  framework for end-to-end aspect-based sentiment triplet extraction,''
  \emph{{IEEE} Trans. Neural Networks Learn. Syst.}, 2021.

\bibitem[Mukherjee et~al.(2021)Mukherjee, Nayak, Butala, Bhattacharya, and
  Goyal]{emnlp21-aste-pointer}
R.~Mukherjee, T.~Nayak, Y.~Butala, S.~Bhattacharya, and P.~Goyal, ``{PASTE:}
  {A} tagging-free decoding framework using pointer networks for aspect
  sentiment triplet extraction,'' in \emph{EMNLP}, 2021, pp. 9279--9291.

\bibitem[Lu et~al.(2022)Lu, Liu, Dai, Xiao, Lin, Han, Sun, and Wu]{acl22-uie}
Y.~Lu, Q.~Liu, D.~Dai, X.~Xiao, H.~Lin, X.~Han, L.~Sun, and H.~Wu, ``Unified
  structure generation for universal information extraction,'' in \emph{ACL},
  2022, pp. 5755--5772.

\bibitem[Wu et~al.(2021{\natexlab{b}})Wu, Xiong, Yi, Yu, Zhu, Gao, and
  Chen]{kbs21-tasd-mejd}
C.~Wu, Q.~Xiong, H.~Yi, Y.~Yu, Q.~Zhu, M.~Gao, and J.~Chen, ``Multiple-element
  joint detection for aspect-based sentiment analysis,'' \emph{Knowl. Based
  Syst.}, vol. 223, p. 107073, 2021.

\bibitem[Cai et~al.(2021)Cai, Xia, and Yu]{acl21-quad-cai}
H.~Cai, R.~Xia, and J.~Yu, ``Aspect-category-opinion-sentiment quadruple
  extraction with implicit aspects and opinions,'' in \emph{ACL-IJCNLP}, 2021,
  pp. 340--350.

\bibitem[Mao et~al.(2022)Mao, Shen, Yang, Zhu, and Cai]{acl22-seq2path}
Y.~Mao, Y.~Shen, J.~Yang, X.~Zhu, and L.~Cai, ``Seq2path: Generating sentiment
  tuples as paths of a tree,'' in \emph{Findings of ACL}, 2022, pp. 2215--2225.

\bibitem[Bao et~al.(2022)Bao, Wang, Jiang, Xiao, and Li]{ijcai22-opinion-tree}
X.~Bao, Z.~Wang, X.~Jiang, R.~Xiao, and S.~Li, ``Aspect-based sentiment
  analysis with opinion tree generation,'' in \emph{IJCAI}, 2022, pp.
  4044--4050.

\bibitem[Sun et~al.(2021)Sun, Liu, Qiu, and Huang]{arxiv21-nlp-paradigm}
T.~Sun, X.~Liu, X.~Qiu, and X.~Huang, ``Paradigm shift in natural language
  processing,'' \emph{CoRR}, vol. abs/2109.12575, 2021.

\bibitem[Kim(2014)]{emnlp14-cnn}
Y.~Kim, ``Convolutional neural networks for sentence classification,'' in
  \emph{EMNLP}, 2014, pp. 1746--1751.

\bibitem[Hochreiter and Schmidhuber(1997)]{nc97-lstm}
S.~Hochreiter and J.~Schmidhuber, ``Long short-term memory,'' \emph{Neural
  computation}, vol.~9, no.~8, pp. 1735--1780, 1997.

\bibitem[Vaswani et~al.(2017)Vaswani, Shazeer, Parmar, Uszkoreit, Jones, Gomez,
  Kaiser, and Polosukhin]{nips17-transformer}
A.~Vaswani, N.~Shazeer, N.~Parmar, J.~Uszkoreit, L.~Jones, A.~N. Gomez,
  {\L}.~Kaiser, and I.~Polosukhin, ``Attention is all you need,'' in
  \emph{NeurIPS}, 2017, pp. 5998--6008.

\bibitem[Lafferty et~al.(2001)Lafferty, McCallum, and Pereira]{icml01-crf}
J.~D. Lafferty, A.~McCallum, and F.~C.~N. Pereira, ``Conditional random fields:
  Probabilistic models for segmenting and labeling sequence data,'' in
  \emph{ICML}, 2001, pp. 282--289.

\bibitem[Sang and Veenstra(1999)]{eacl99-tagging-scheme}
E.~F. T.~K. Sang and J.~Veenstra, ``Representing text chunks,'' in \emph{EACL},
  1999, pp. 173--179.

\bibitem[Chen(2018)]{thesis18-mrc-cdq}
D.~Chen, ``Neural reading comprehension and beyond,'' Ph.D. dissertation,
  Stanford University, 2018.

\bibitem[Sutskever et~al.(2014)Sutskever, Vinyals, and Le]{nips14-seq2seq}
I.~Sutskever, O.~Vinyals, and Q.~V. Le, ``Sequence to sequence learning with
  neural networks,'' in \emph{NeurIPS}, 2014, pp. 3104--3112.

\bibitem[Pontiki et~al.(2014)Pontiki, Galanis, Pavlopoulos, Papageorgiou,
  Androutsopoulos, and Manandhar]{semeval14-absa}
M.~Pontiki, D.~Galanis, J.~Pavlopoulos, H.~Papageorgiou, I.~Androutsopoulos,
  and S.~Manandhar, ``Semeval-2014 task 4: Aspect based sentiment analysis,''
  in \emph{SemEval@COLING}, P.~Nakov and T.~Zesch, Eds., 2014, pp. 27--35.

\bibitem[Pontiki et~al.(2015)Pontiki, Galanis, Papageorgiou, Manandhar, and
  Androutsopoulos]{semeval15-absa}
M.~Pontiki, D.~Galanis, H.~Papageorgiou, S.~Manandhar, and I.~Androutsopoulos,
  ``Semeval-2015 task 12: Aspect based sentiment analysis,'' in
  \emph{SemEval@NAACL-HLT}, D.~M. Cer, D.~Jurgens, P.~Nakov, and T.~Zesch,
  Eds., 2015, pp. 486--495.

\bibitem[Pontiki et~al.(2016)Pontiki, Galanis, Papageorgiou, Androutsopoulos,
  Manandhar, Al{-}Smadi, Al{-}Ayyoub, Zhao, Qin, Clercq, Hoste, Apidianaki,
  Tannier, Loukachevitch, Kotelnikov, Bel, Zafra, and Eryigit]{semeval16-absa}
M.~Pontiki, D.~Galanis, H.~Papageorgiou, I.~Androutsopoulos, S.~Manandhar,
  M.~Al{-}Smadi, M.~Al{-}Ayyoub, Y.~Zhao, B.~Qin, O.~D. Clercq, V.~Hoste,
  M.~Apidianaki, X.~Tannier, N.~V. Loukachevitch, E.~V. Kotelnikov, N.~Bel,
  S.~M.~J. Zafra, and G.~Eryigit, ``Semeval-2016 task 5: Aspect based sentiment
  analysis,'' in \emph{SemEval@NAACL-HLT}, 2016, pp. 19--30.

\bibitem[Wang et~al.(2019)Wang, Sun, Li, Liu, Si, Zhang, and
  Zhou]{acl19-absa-qa}
J.~Wang, C.~Sun, S.~Li, X.~Liu, L.~Si, M.~Zhang, and G.~Zhou, ``Aspect
  sentiment classification towards question-answering with reinforced
  bidirectional attention network,'' in \emph{ACL}, 2019, pp. 3548--3557.

\bibitem[Jiang et~al.(2019)Jiang, Chen, Xu, Ao, and Yang]{emnlp19-absa-mams}
Q.~Jiang, L.~Chen, R.~Xu, X.~Ao, and M.~Yang, ``A challenge dataset and
  effective models for aspect-based sentiment analysis,'' in
  \emph{EMNLP-IJCNLP}, 2019, pp. 6280--6285.

\bibitem[Xing et~al.(2020)Xing, Jin, Jin, Wang, Zhang, and
  Huang]{emnlp20-asc-tasty}
X.~Xing, Z.~Jin, D.~Jin, B.~Wang, Q.~Zhang, and X.~Huang, ``Tasty burgers,
  soggy fries: Probing aspect robustness in aspect-based sentiment analysis,''
  in \emph{EMNLP}, 2020, pp. 3594--3605.

\bibitem[Bu et~al.(2021)Bu, Ren, Zheng, Yang, Wang, Zhang, and
  Wu]{naacl21-acsc-asap}
J.~Bu, L.~Ren, S.~Zheng, Y.~Yang, J.~Wang, F.~Zhang, and W.~Wu, ``{ASAP:} {A}
  chinese review dataset towards aspect category sentiment analysis and rating
  prediction,'' in \emph{NAACL-HLT}, 2021, pp. 2069--2079.

\bibitem[Zhang et~al.(2018)Zhang, Wang, and Liu]{widm18-sa-survey}
L.~Zhang, S.~Wang, and B.~Liu, ``Deep learning for sentiment analysis: {A}
  survey,'' \emph{Wiley Interdiscip. Rev. Data Min. Knowl. Discov.}, vol.~8,
  no.~4, 2018.

\bibitem[Schouten et~al.(2018)Schouten, van~der Weijde, Frasincar, and
  Dekker]{tycb18-acd}
K.~Schouten, O.~van~der Weijde, F.~Frasincar, and R.~Dekker, ``Supervised and
  unsupervised aspect category detection for sentiment analysis with
  co-occurrence data,'' \emph{{IEEE} Trans. Cybern.}, vol.~48, no.~4, pp.
  1263--1275, 2018.

\bibitem[Wang and Pan(2018{\natexlab{a}})]{acl18-ate-recursive}
W.~Wang and S.~J. Pan, ``Recursive neural structural correspondence network for
  cross-domain aspect and opinion co-extraction,'' in \emph{ACL}, 2018, pp.
  2171--2181.

\bibitem[Wu et~al.(2020{\natexlab{c}})Wu, Zhao, Dai, Huang, and
  Chen]{aaai20-towe-transfer}
Z.~Wu, F.~Zhao, X.~Dai, S.~Huang, and J.~Chen, ``Latent opinions transfer
  network for target-oriented opinion words extraction,'' in \emph{AAAI}, 2020,
  pp. 9298--9305.

\bibitem[Kipf and Welling(2017)]{iclr17-gcn}
T.~N. Kipf and M.~Welling, ``Semi-supervised classification with graph
  convolutional networks,'' in \emph{ICLR}, 2017.

\bibitem[Tay et~al.(2018)Tay, Tuan, and Hui]{aaai18-asc-yitay}
Y.~Tay, L.~A. Tuan, and S.~C. Hui, ``Learning to attend via word-aspect
  associative fusion for aspect-based sentiment analysis,'' in \emph{AAAI},
  2018, pp. 5956--5963.

\bibitem[Xing et~al.(2019)Xing, Liao, Song, Wang, Zhang, Wang, and
  Huang]{ijcai19-asc-early}
B.~Xing, L.~Liao, D.~Song, J.~Wang, F.~Zhang, Z.~Wang, and H.~Huang, ``Earlier
  attention? aspect-aware {LSTM} for aspect-based sentiment analysis,'' in
  \emph{IJCAI}, 2019, pp. 5313--5319.

\bibitem[He et~al.(2018)He, Lee, Ng, and Dahlmeier]{coling18-asc-ruidan}
R.~He, W.~S. Lee, H.~T. Ng, and D.~Dahlmeier, ``Effective attention modeling
  for aspect-level sentiment classification,'' in \emph{COLING}, 2018, pp.
  1121--1131.

\bibitem[Zhang et~al.(2016)Zhang, Zhang, and Vo]{aaai16-asc-gated}
M.~Zhang, Y.~Zhang, and D.~Vo, ``Gated neural networks for targeted sentiment
  analysis,'' in \emph{AAAI}, 2016, pp. 3087--3093.

\bibitem[Sun et~al.(2019{\natexlab{b}})Sun, Huang, and Qiu]{naacl19-utilizing}
C.~Sun, L.~Huang, and X.~Qiu, ``Utilizing {BERT} for aspect-based sentiment
  analysis via constructing auxiliary sentence,'' in \emph{NAACL-HLT}, 2019,
  pp. 380--385.

\bibitem[Wu and Ong(2021)]{aaai21-asc-context-bert}
Z.~Wu and D.~C. Ong, ``Context-guided {BERT} for targeted aspect-based
  sentiment analysis,'' in \emph{AAAI}, 2021, pp. 14\,094--14\,102.

\bibitem[Dai et~al.(2021)Dai, Yan, Sun, Liu, and Qiu]{naacl21-asc-roberta}
J.~Dai, H.~Yan, T.~Sun, P.~Liu, and X.~Qiu, ``Does syntax matter? {A} strong
  baseline for aspect-based sentiment analysis with roberta,'' in
  \emph{NAACL-HLT}, 2021, pp. 1816--1829.

\bibitem[Brun et~al.(2014)Brun, Popa, and Roux]{semeval14-xrce}
C.~Brun, D.~N. Popa, and C.~Roux, ``{XRCE:} hybrid classification for
  aspect-based sentiment analysis,'' in \emph{SemEval@COLING}, 2014, pp.
  838--842.

\bibitem[Kiritchenko et~al.(2014)Kiritchenko, Zhu, Cherry, and
  Mohammad]{semeval14-nrc}
S.~Kiritchenko, X.~Zhu, C.~Cherry, and S.~Mohammad, ``Nrc-canada-2014:
  Detecting aspects and sentiment in customer reviews,'' in
  \emph{SemEval@COLING}, 2014, pp. 437--442.

\bibitem[Zhou et~al.(2020)Zhou, Cui, Hu, Zhang, Yang, Liu, Wang, Li, and
  Sun]{aiopen20-gnn-survey}
J.~Zhou, G.~Cui, S.~Hu, Z.~Zhang, C.~Yang, Z.~Liu, L.~Wang, C.~Li, and M.~Sun,
  ``Graph neural networks: {A} review of methods and applications,'' \emph{{AI}
  Open}, vol.~1, pp. 57--81, 2020.

\bibitem[Huang and Carley(2019)]{emnlp19-syntax-gat}
B.~Huang and K.~Carley, ``Syntax-aware aspect level sentiment classification
  with graph attention networks,'' in \emph{EMNLP-IJCNLP}, 2019, pp.
  5469--5477.

\bibitem[Zhang and Qian(2020)]{emnlp20-bigcn}
M.~Zhang and T.~Qian, ``Convolution over hierarchical syntactic and lexical
  graphs for aspect level sentiment analysis,'' in \emph{EMNLP}, 2020, pp.
  3540--3549.

\bibitem[Wang et~al.(2020{\natexlab{a}})Wang, Shen, Yang, Quan, and
  Wang]{acl20-relational}
K.~Wang, W.~Shen, Y.~Yang, X.~Quan, and R.~Wang, ``Relational graph attention
  network for aspect-based sentiment analysis,'' in \emph{ACL}, 2020, pp.
  3229--3238.

\bibitem[Ruder et~al.(2016)Ruder, Ghaffari, and Breslin]{emnlp16-acsc-ruder}
S.~Ruder, P.~Ghaffari, and J.~G. Breslin, ``A hierarchical model of reviews for
  aspect-based sentiment analysis,'' in \emph{EMNLP}, 2016, pp. 999--1005.

\bibitem[Chen et~al.(2020{\natexlab{c}})Chen, Sun, Wang, Li, Si, Zhang, and
  Zhou]{acl20-asc-document}
X.~Chen, C.~Sun, J.~Wang, S.~Li, L.~Si, M.~Zhang, and G.~Zhou, ``Aspect
  sentiment classification with document-level sentiment preference modeling,''
  in \emph{ACL}, 2020, pp. 3667--3677.

\bibitem[Zhang et~al.(2021{\natexlab{c}})Zhang, Deng, Li, Bing, and
  Lam]{emnlp21-findings-uabsa-qa}
W.~Zhang, Y.~Deng, X.~Li, L.~Bing, and W.~Lam, ``Aspect-based sentiment
  analysis in question answering forums,'' in \emph{Findings of EMNLP}, 2021,
  pp. 4582--4591.

\bibitem[Hu et~al.(2019{\natexlab{b}})Hu, Zhao, Zhang, Cai, Su, Cheng, and
  Shen]{emnlp19-can}
M.~Hu, S.~Zhao, L.~Zhang, K.~Cai, Z.~Su, R.~Cheng, and X.~Shen, ``{CAN:}
  constrained attention networks for multi-aspect sentiment analysis,'' in
  \emph{EMNLP-IJCNLP}, 2019, pp. 4600--4609.

\bibitem[Ma et~al.(2018)Ma, Peng, and Cambria]{aaai18-sentilstm}
Y.~Ma, H.~Peng, and E.~Cambria, ``Targeted aspect-based sentiment analysis via
  embedding commonsense knowledge into an attentive {LSTM},'' in \emph{AAAI},
  2018, pp. 5876--5883.

\bibitem[Dai et~al.(2020)Dai, Peng, Chen, and Ding]{emnlp20-acsa-incremental}
Z.~Dai, C.~Peng, H.~Chen, and Y.~Ding, ``A multi-task incremental learning
  framework with category name embedding for aspect-category sentiment
  analysis,'' in \emph{EMNLP}, 2020, pp. 6955--6965.

\bibitem[Li et~al.(2020{\natexlab{b}})Li, Yang, Yin, Pan, Cui, Huang, and
  Wei]{ccl20-acsa}
Y.~Li, Z.~Yang, C.~Yin, X.~Pan, L.~Cui, Q.~Huang, and T.~Wei, ``A joint model
  for aspect-category sentiment analysis with shared sentiment prediction
  layer,'' in \emph{CCL}, 2020, pp. 388--400.

\bibitem[Hsu et~al.(2021)Hsu, Chen, Huang, and Chen]{emnlp21-absa-spdaug}
T.-W. Hsu, C.-C. Chen, H.-H. Huang, and H.-H. Chen, ``Semantics-preserved data
  augmentation for aspect-based sentiment analysis,'' in \emph{EMNLP}, 2021,
  pp. 4417--4422.

\bibitem[Brun and Nikoulina(2018)]{wassa18-tasd-wild}
C.~Brun and V.~Nikoulina, ``Aspect based sentiment analysis into the wild,'' in
  \emph{WASSA@EMNLP}, 2018, pp. 116--122.

\bibitem[Mikolov et~al.(2013)Mikolov, Sutskever, Chen, Corrado, and
  Dean]{nips13-word2vec}
T.~Mikolov, I.~Sutskever, K.~Chen, G.~S. Corrado, and J.~Dean, ``Distributed
  representations of words and phrases and their compositionality,'' in
  \emph{NeurIPS}, 2013, pp. 3111--3119.

\bibitem[Pennington et~al.(2014)Pennington, Socher, and Manning]{emnlp14-glove}
J.~Pennington, R.~Socher, and C.~Manning, ``{G}lo{V}e: Global vectors for word
  representation,'' in \emph{EMNLP}, 2014, pp. 1532--1543.

\bibitem[Song et~al.(2019)Song, Wang, Jiang, Liu, and Rao]{arxiv19-attentional}
Y.~Song, J.~Wang, T.~Jiang, Z.~Liu, and Y.~Rao, ``Attentional encoder network
  for targeted sentiment classification,'' \emph{arXiv}, 2019.

\bibitem[Rietzler et~al.(2020)Rietzler, Stabinger, Opitz, and
  Engl]{lrec20-adapt}
A.~Rietzler, S.~Stabinger, P.~Opitz, and S.~Engl, ``Adapt or get left behind:
  Domain adaptation through {BERT} language model finetuning for aspect-target
  sentiment classification,'' in \emph{LREC}, 2020, pp. 4933--4941.

\bibitem[Lewis et~al.(2020)Lewis, Liu, Goyal, Ghazvininejad, Mohamed, Levy,
  Stoyanov, and Zettlemoyer]{acl20-bart}
M.~Lewis, Y.~Liu, N.~Goyal, M.~Ghazvininejad, A.~Mohamed, O.~Levy, V.~Stoyanov,
  and L.~Zettlemoyer, ``{BART:} denoising sequence-to-sequence pre-training for
  natural language generation, translation, and comprehension,'' in \emph{ACL},
  2020, pp. 7871--7880.

\bibitem[Raffel et~al.(2020)Raffel, Shazeer, Roberts, Lee, Narang, Matena,
  Zhou, Li, and Liu]{jmlr20-t5}
C.~Raffel, N.~Shazeer, A.~Roberts, K.~Lee, S.~Narang, M.~Matena, Y.~Zhou,
  W.~Li, and P.~J. Liu, ``Exploring the limits of transfer learning with a
  unified text-to-text transformer,'' \emph{J. Mach. Learn. Res.}, vol.~21, pp.
  140:1--140:67, 2020.

\bibitem[Wu et~al.(2020{\natexlab{d}})Wu, Chen, Kao, and
  Liu]{acl20-asc-perturbed}
Z.~Wu, Y.~Chen, B.~Kao, and Q.~Liu, ``Perturbed masking: Parameter-free probing
  for analyzing and interpreting {BERT},'' in \emph{ACL}, 2020, pp. 4166--4176.

\bibitem[Qiu et~al.(2020)Qiu, Sun, Xu, Shao, Dai, and Huang]{plm-survey-qxp}
X.~Qiu, T.~Sun, Y.~Xu, Y.~Shao, N.~Dai, and X.~Huang, ``Pre-trained models for
  natural language processing: {A} survey,'' \emph{CoRR}, vol. abs/2003.08271,
  2020.

\bibitem[Blitzer et~al.(2006)Blitzer, McDonald, and Pereira]{emnlp06-domain}
J.~Blitzer, R.~McDonald, and F.~Pereira, ``Domain adaptation with structural
  correspondence learning,'' in \emph{EMNLP}, 2006, pp. 120--128.

\bibitem[Jakob and Gurevych(2010)]{emnlp10-ate-extracting}
N.~Jakob and I.~Gurevych, ``Extracting opinion targets in a single and
  cross-domain setting with conditional random fields,'' in \emph{EMNLP}, 2010,
  pp. 1035--1045.

\bibitem[Chernyshevich(2014)]{semeval14-ate-ihs}
M.~Chernyshevich, ``{IHS} {R}{\&}{D} {B}elarus: Cross-domain extraction of
  product features using {CRF},'' in \emph{{S}em{E}val}, 2014, pp. 309--313.

\bibitem[Wang and Pan(2019{\natexlab{a}})]{cl19-ate-syntactically}
W.~Wang and S.~J. Pan, ``Syntactically meaningful and transferable recursive
  neural networks for aspect and opinion extraction,'' \emph{Computational
  Linguistics}, vol.~45, no.~4, pp. 705--736, 2019.

\bibitem[Yang et~al.(2019)Yang, Yin, Qu, Tu, Shen, and Chen]{tac19-neural}
M.~Yang, W.~Yin, Q.~Qu, W.~Tu, Y.~Shen, and X.~Chen, ``Neural attentive network
  for cross-domain aspect-level sentiment classification,'' \emph{{IEEE} Trans.
  Affect. Comput.}, 2019.

\bibitem[Zhang et~al.(2021{\natexlab{d}})Zhang, Liu, Qian, Xiang, Cui, Zhou,
  and Chen]{tkde21-asc-eatn}
K.~Zhang, Q.~Liu, H.~Qian, B.~Xiang, Q.~Cui, J.~Zhou, and E.~Chen, ``Eatn: An
  efficient adaptive transfer network for aspect-level sentiment analysis,''
  \emph{IEEE TKDE}, 2021.

\bibitem[Wang and Pan(2019{\natexlab{b}})]{aaai19-ate-transferable}
W.~Wang and S.~J. Pan, ``Transferable interactive memory network for domain
  adaptation in fine-grained opinion extraction,'' in \emph{AAAI}, 2019, pp.
  7192--7199.

\bibitem[Li et~al.(2019{\natexlab{c}})Li, Li, Wei, Bing, Zhang, and
  Yang]{emnlp19-uabsa-transferable}
Z.~Li, X.~Li, Y.~Wei, L.~Bing, Y.~Zhang, and Q.~Yang, ``Transferable end-to-end
  aspect-based sentiment analysis with selective adversarial learning,'' in
  \emph{EMNLP-IJCNLP}, 2019, pp. 4590--4600.

\bibitem[Chen and Qian(2021)]{acl21-ate-bridge}
Z.~Chen and T.~Qian, ``Bridge-based active domain adaptation for aspect term
  extraction,'' in \emph{ACL-IJCNLP}, 2021, pp. 317--327.

\bibitem[Liang et~al.(2021{\natexlab{b}})Liang, Wang, and
  Lv]{tnnls21-ate-weakly}
T.~Liang, W.~Wang, and F.~Lv, ``Weakly supervised domain adaptation for aspect
  extraction via multilevel interaction transfer,'' \emph{IEEE Transactions on
  Neural Networks and Learning Systems}, 2021.

\bibitem[Yu et~al.(2021{\natexlab{b}})Yu, Gong, and Xia]{acl21-uabsa-cross}
J.~Yu, C.~Gong, and R.~Xia, ``Cross-domain review generation for aspect-based
  sentiment analysis,'' in \emph{Findings of ACL-IJCNLP}, 2021, pp. 4767--4777.

\bibitem[Li et~al.(2022)Li, Yu, and Xia]{naacl22-xdomain}
J.~Li, J.~Yu, and R.~Xia, ``Generative cross-domain data augmentation for
  aspect and opinion co-extraction,'' in \emph{NAACL}, 2022, pp. 4219--4229.

\bibitem[Ding et~al.(2017)Ding, Yu, and Jiang]{aaai17-ate-recurrent}
Y.~Ding, J.~Yu, and J.~Jiang, ``Recurrent neural networks with auxiliary labels
  for cross-domain opinion target extraction,'' in \emph{AAAI}, 2017, pp.
  3436--3442.

\bibitem[Li et~al.(2012)Li, Pan, Jin, Yang, and Zhu]{acl12-ate-cross}
F.~Li, S.~J. Pan, O.~Jin, Q.~Yang, and X.~Zhu, ``Cross-domain co-extraction of
  sentiment and topic lexicons,'' in \emph{ACL}, 2012, pp. 410--419.

\bibitem[Gong et~al.(2020)Gong, Yu, and Xia]{emnlp20-uabsa-uf}
C.~Gong, J.~Yu, and R.~Xia, ``Unified feature and instance based domain
  adaptation for aspect-based sentiment analysis,'' in \emph{EMNLP}, 2020, pp.
  7035--7045.

\bibitem[Pereg et~al.(2020)Pereg, Korat, and
  Wasserblat]{coling20-ate-syntactically}
O.~Pereg, D.~Korat, and M.~Wasserblat, ``Syntactically aware cross-domain
  aspect and opinion terms extraction,'' in \emph{COLING}, 2020, pp.
  1772--1777.

\bibitem[Xu et~al.(2020{\natexlab{c}})Xu, Liu, Shu, and
  Yu]{emnlp20-uabsa-dombert}
H.~Xu, B.~Liu, L.~Shu, and P.~Yu, ``{D}om{BERT}: Domain-oriented language model
  for aspect-based sentiment analysis,'' in \emph{Findings of EMNLP}, 2020, pp.
  1725--1731.

\bibitem[Zhou et~al.(2015{\natexlab{b}})Zhou, Wan, and Xiao]{taslp15-xabsa-wxj}
X.~Zhou, X.~Wan, and J.~Xiao, ``Clopinionminer: Opinion target extraction in a
  cross-language scenario,'' \emph{{IEEE} {ACM} Trans. Audio Speech Lang.
  Process.}, vol.~23, no.~4, pp. 619--630, 2015.

\bibitem[Klinger and Cimiano(2015)]{conll15-xabsa-instance}
R.~Klinger and P.~Cimiano, ``Instance selection improves cross-lingual model
  training for fine-grained sentiment analysis,'' in \emph{CoNLL}, 2015, pp.
  153--163.

\bibitem[Wang and Pan(2018{\natexlab{b}})]{ijcai18-xabsa-transition}
W.~Wang and S.~J. Pan, ``Transition-based adversarial network for cross-lingual
  aspect extraction,'' in \emph{IJCAI}, 2018, pp. 4475--4481.

\bibitem[Jebbara and Cimiano(2019)]{naacl19-xabsa-zs-ate}
S.~Jebbara and P.~Cimiano, ``Zero-shot cross-lingual opinion target
  extraction,'' in \emph{NAACL}, 2019, pp. 2486--2495.

\bibitem[Lambert(2015)]{acl15-xabsa-smt}
P.~Lambert, ``Aspect-level cross-lingual sentiment classification with
  constrained {SMT},'' in \emph{ACL}, 2015, pp. 781--787.

\bibitem[Barnes et~al.(2016)Barnes, Lambert, and
  Badia]{coling16-xabsa-bilingual-embed}
J.~Barnes, P.~Lambert, and T.~Badia, ``Exploring distributional representations
  and machine translation for aspect-based cross-lingual sentiment
  classification,'' in \emph{COLING}, 2016, pp. 1613--1623.

\bibitem[Akhtar et~al.(2018)Akhtar, Sawant, Sen, Ekbal, and
  Bhattacharyya]{naacl18-xabsa-sparse}
M.~S. Akhtar, P.~Sawant, S.~Sen, A.~Ekbal, and P.~Bhattacharyya, ``Solving data
  sparsity for aspect based sentiment analysis using cross-linguality and
  multi-linguality,'' in \emph{NAACL}, 2018, pp. 572--582.

\bibitem[Li et~al.(2020{\natexlab{c}})Li, Bing, Zhang, Li, and
  Lam]{arxiv20-xabsa-lx}
X.~Li, L.~Bing, W.~Zhang, Z.~Li, and W.~Lam, ``Unsupervised cross-lingual
  adaptation for sequence tagging and beyond,'' \emph{CoRR}, vol.
  abs/2010.12405, 2020.

\bibitem[Zhang et~al.(2021{\natexlab{e}})Zhang, He, Peng, Bing, and
  Lam]{emnlp21-xabsa-wxzhang}
W.~Zhang, R.~He, H.~Peng, L.~Bing, and W.~Lam, ``Cross-lingual aspect-based
  sentiment analysis with aspect term code-switching,'' in \emph{EMNLP}, 2021,
  pp. 9220--9230.

\bibitem[Dyer et~al.(2013)Dyer, Chahuneau, and Smith]{fastalign}
C.~Dyer, V.~Chahuneau, and N.~A. Smith, ``A simple, fast, and effective
  reparameterization of {IBM} model 2,'' in \emph{NAACL}, 2013, pp. 644--648.

\bibitem[Conneau et~al.(2020)Conneau, Khandelwal, Goyal, Chaudhary, Wenzek,
  Guzm{\'a}n, Grave, Ott, Zettlemoyer, and Stoyanov]{acl20-xlmr}
A.~Conneau, K.~Khandelwal, N.~Goyal, V.~Chaudhary, G.~Wenzek, F.~Guzm{\'a}n,
  E.~Grave, M.~Ott, L.~Zettlemoyer, and V.~Stoyanov, ``Unsupervised
  cross-lingual representation learning at scale,'' in \emph{ACL}, 2020, pp.
  8440--8451.

\bibitem[Pires et~al.(2019)Pires, Schlinger, and Garrette]{acl19-mbert-study}
T.~Pires, E.~Schlinger, and D.~Garrette, ``How multilingual is multilingual
  bert?'' in \emph{ACL}, 2019, pp. 4996--5001.

\bibitem[K et~al.(2020)K, Wang, Mayhew, and Roth]{iclr20-mbert-study}
K.~K, Z.~Wang, S.~Mayhew, and D.~Roth, ``Cross-lingual ability of multilingual
  {BERT:} an empirical study,'' in \emph{ICLR}, 2020.

\bibitem[Wang et~al.(2020{\natexlab{b}})Wang, Wang, Sun, Li, Liu, Si, Zhang,
  and Zhou]{aaai20-sentiment-dialog}
J.~Wang, J.~Wang, C.~Sun, S.~Li, X.~Liu, L.~Si, M.~Zhang, and G.~Zhou,
  ``Sentiment classification in customer service dialogue with topic-aware
  multi-task learning,'' in \emph{AAAI}, 2020, pp. 9177--9184.

\bibitem[Yu et~al.(2022)Yu, Wang, Xia, and Li]{ijcai22-mmabsa}
J.~Yu, J.~Wang, R.~Xia, and J.~Li, ``Targeted multimodal sentiment
  classification based on coarse-to-fine grained image-target matching,'' in
  \emph{IJCAI}, 2022, pp. 4482--4488.

\bibitem[Ling et~al.(2022)Ling, Yu, and Xia]{acl22-mmabsa}
Y.~Ling, J.~Yu, and R.~Xia, ``Vision-language pre-training for multimodal
  aspect-based sentiment analysis,'' in \emph{ACL}, 2022, pp. 2149--2159.

\bibitem[Wu et~al.(2020{\natexlab{e}})Wu, Cheng, Wang, Li, and
  Chi]{nlpcc20-mm-ate}
H.~Wu, S.~Cheng, J.~Wang, S.~Li, and L.~Chi, ``Multimodal aspect extraction
  with region-aware alignment network,'' in \emph{NLPCC}, 2020, pp. 145--156.

\bibitem[Zhou et~al.(2021{\natexlab{b}})Zhou, Zhao, Huang, Hu, and
  He]{neuralcomp21-mm-acsc}
J.~Zhou, J.~Zhao, J.~X. Huang, Q.~V. Hu, and L.~He, ``{MASAD:} {A} large-scale
  dataset for multimodal aspect-based sentiment analysis,''
  \emph{Neurocomputing}, vol. 455, pp. 47--58, 2021.

\bibitem[Xu et~al.(2019{\natexlab{b}})Xu, Mao, and Chen]{aaai19-mm-acsc}
N.~Xu, W.~Mao, and G.~Chen, ``Multi-interactive memory network for aspect based
  multimodal sentiment analysis,'' in \emph{AAAI}, 2019, pp. 371--378.

\bibitem[Yu and Jiang(2019)]{ijcai19-mm-atsc}
J.~Yu and J.~Jiang, ``Adapting {BERT} for target-oriented multimodal sentiment
  classification,'' in \emph{IJCAI}, 2019, pp. 5408--5414.

\bibitem[Yu et~al.(2020)Yu, Jiang, and Xia]{taslp20-mm-atsc}
J.~Yu, J.~Jiang, and R.~Xia, ``Entity-sensitive attention and fusion network
  for entity-level multimodal sentiment classification,'' \emph{{IEEE} {ACM}
  Trans. Audio Speech Lang. Process.}, vol.~28, pp. 429--439, 2020.

\bibitem[Khan and Fu(2021)]{mm21-mm-atsc}
Z.~Khan and Y.~Fu, ``Exploiting {BERT} for multimodal target sentiment
  classification through input space translation,'' in \emph{{MM}}, 2021, pp.
  3034--3042.

\bibitem[Ju et~al.(2021)Ju, Zhang, Xiao, Li, Li, Zhang, and
  Zhou]{emnlp21-mm-e2e-absa}
X.~Ju, D.~Zhang, R.~Xiao, J.~Li, S.~Li, M.~Zhang, and G.~Zhou, ``Joint
  multi-modal aspect-sentiment analysis with auxiliary cross-modal relation
  detection,'' in \emph{EMNLP}, 2021, pp. 4395--4405.

\bibitem[Delange et~al.(2021)Delange, Aljundi, Masana, Parisot, Jia, Leonardis,
  Slabaugh, and Tuytelaars]{delange2021continual}
M.~Delange, R.~Aljundi, M.~Masana, S.~Parisot, X.~Jia, A.~Leonardis,
  G.~Slabaugh, and T.~Tuytelaars, ``A continual learning survey: Defying
  forgetting in classification tasks,'' \emph{IEEE TPAMI}, 2021.

\bibitem[Chen et~al.(2015)Chen, Ma, and Liu]{acl15-lifelongsa}
Z.~Chen, N.~Ma, and B.~Liu, ``Lifelong learning for sentiment classification,''
  in \emph{ACL}, 2015, pp. 750--756.

\bibitem[Wang et~al.(2018{\natexlab{b}})Wang, Lv, Mazumder, Fei, and
  Liu]{bigdata18-lifelongabsa}
S.~Wang, G.~Lv, S.~Mazumder, G.~Fei, and B.~Liu, ``Lifelong learning memory
  networks for aspect sentiment classification,'' in \emph{IEEE BigData}, 2018,
  pp. 861--870.

\bibitem[Geng et~al.(2021)Geng, Yang, Yuan, Wang, Ao, and
  Xu]{sigir21-lifelongsa}
B.~Geng, M.~Yang, F.~Yuan, S.~Wang, X.~Ao, and R.~Xu, ``Iterative network
  pruning with uncertainty regularization for lifelong sentiment
  classification,'' in \emph{SIGIR}, 2021, pp. 1229--1238.

\bibitem[Wang et~al.(2020{\natexlab{c}})Wang, Wang, Mazumder, Liu, Yang, and
  Li]{coling20-lifelongsa}
H.~Wang, S.~Wang, S.~Mazumder, B.~Liu, Y.~Yang, and T.~Li, ``Bayes-enhanced
  lifelong attention networks for sentiment classification,'' in \emph{COLING},
  2020, pp. 580--591.

\bibitem[Ke et~al.(2021{\natexlab{a}})Ke, Liu, Xu, and
  Shu]{emnlp21-continualabsa}
Z.~Ke, B.~Liu, H.~Xu, and L.~Shu, ``{CLASSIC:} continual and contrastive
  learning of aspect sentiment classification tasks,'' in \emph{EMNLP}, 2021,
  pp. 6871--6883.

\bibitem[Ke et~al.(2021{\natexlab{b}})Ke, Xu, and Liu]{naacl21-continualabsa}
Z.~Ke, H.~Xu, and B.~Liu, ``Adapting {BERT} for continual learning of a
  sequence of aspect sentiment classification tasks,'' in \emph{NAACL-HLT},
  2021, pp. 4746--4755.

\end{thebibliography}

\end{document}